\pdfoutput=1

\documentclass[11pt]{article}

\usepackage[preprint]{acl}

\usepackage{times}
\usepackage{latexsym}

\usepackage[T1]{fontenc}

\usepackage[utf8]{inputenc}

\usepackage[whole]{bxcjkjatype}

\usepackage{microtype}

\usepackage{inconsolata}

\usepackage{graphicx}

\usepackage{amsmath}
\usepackage{booktabs}
\usepackage{cleveref} 

\usepackage{amsmath}
\usepackage{pifont}
\usepackage{array}

\usepackage{amssymb}
\usepackage{pifont}

\usepackage{listings}
\begin{document}
\begin{CJK}{UTF8}{min}
\title{A Dataset for Evaluating LLM-based Evaluation Functions for Research Question Extraction Task}




\author{
 \textbf{Yuya Fujisaki \textsuperscript{1}},
 \textbf{Shiro Takagi\textsuperscript{2}},
 \textbf{Hideki Asoh \textsuperscript{3}},
 \textbf{Wataru Kumagai \textsuperscript{4}},
\\
\\
 \textsuperscript{1}Japan Advanced Institute of Science and Technology, www.yuya.fujisaki@gmail.com \\
 \textsuperscript{2}Independent Researcher, takagi4646@gmail.com \\
 \textsuperscript{3}National Institute of Advanced Industrial Science and Technology, h.asoh@aist.go.jp \\
 \textsuperscript{4}OMRON SINIC X Corporation, wataru.kumagai@sinicx.com 
}

\maketitle
\begin{abstract}
The progress in text summarization techniques has been remarkable. However the task of accurately extracting and summarizing necessary information from highly specialized documents such as research papers has not been sufficiently investigated. We are focusing on the task of extracting research questions (RQ) from research papers and construct a new dataset consisting of machine learning papers, RQ extracted from these papers by GPT-4, and human evaluations of the extracted RQ from multiple perspectives. Using this dataset, we systematically compared recently proposed LLM-based evaluation functions for summarizations, and found that none of the functions showed sufficiently high correlations with human evaluations. We expect our dataset provides a foundation for further research on developing better evaluation functions tailored to the RQ extraction task, and contribute to enhance the performance of the task. The dataset is available at \href{https://github.com/auto-res/PaperRQ-HumanAnno-Dataset}{PaperRQ-HumanAnno-Dataset}.
\end{abstract}
  
\section{Introduction}
\begin{figure*}[t]
  \centering
  \includegraphics[width=0.98\linewidth]{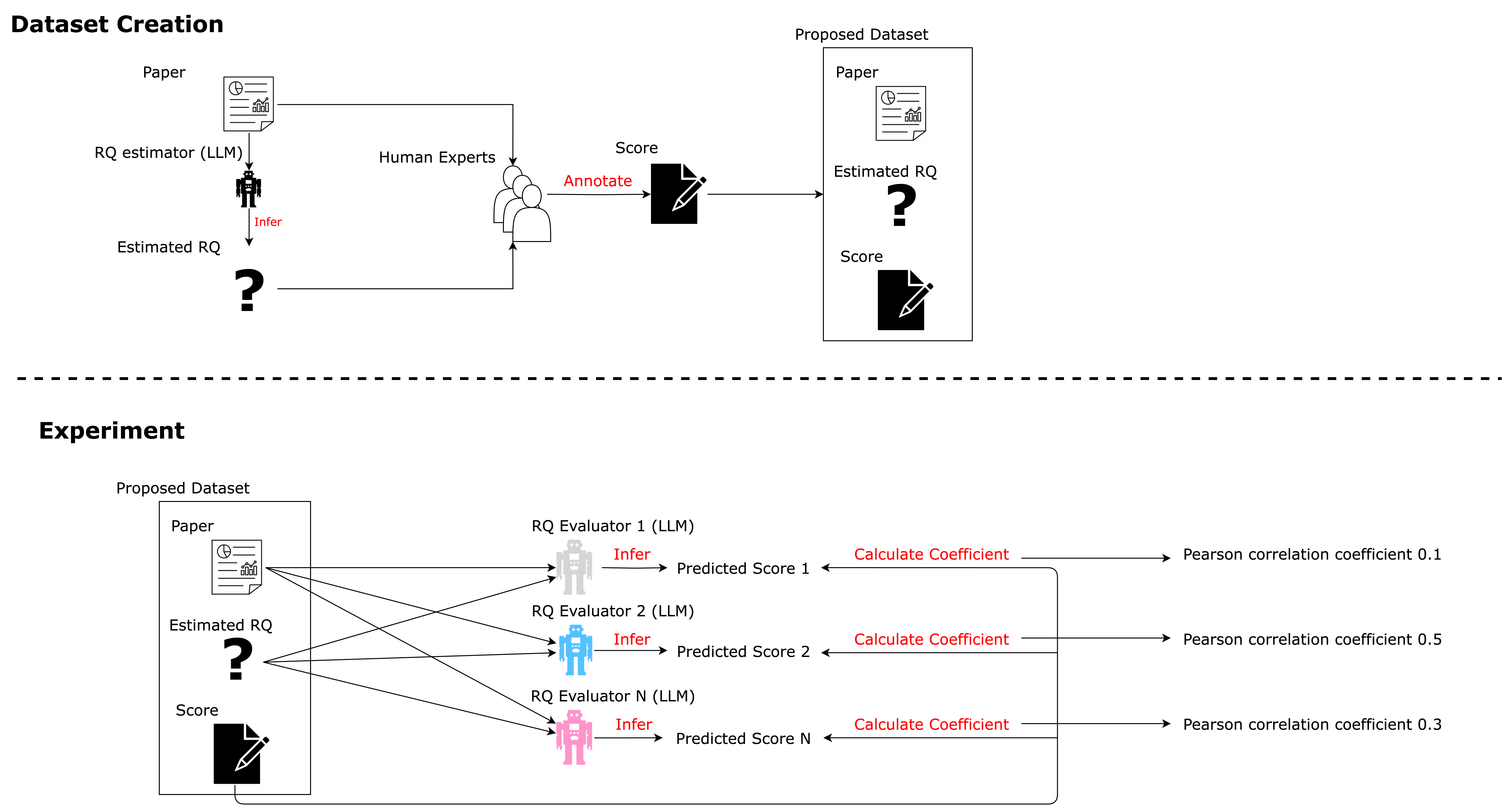}
  \caption{This study has two main processes. First, we constructed a dataset consisting of papers, RQ extracted by an LLM, and human evaluation scores of the RQ quality based on the paper abstract and introduction. Second, using this dataset, we analyzed the correlation between the output scores of various LLM-based evaluation functions and human scores, and identified the evaluation function that is closest to human judgment. Through this series of processes, we confirmed the effectiveness of automatic evaluation of RQ using LLM.}
  \label{fig:image-of-whole-process}
\end{figure*}

To understand research papers, it's crucial to accurately grasp the Research Question (RQ). The RQ is the specific question set by the authors to address a particular research problem. It guides the research direction and narrows the focus of investigations and experiments. Proper understanding of the RQ is essential for clarifying the research purpose and scope and comprehending the paper's main arguments.

However, research papers tend to have a complex structure, use many technical terms, and have important information scattered throughout the document, making RQ hard to grasp easily.

Considering these characteristics of research papers, automatic RQ extraction, which involves identifying the key components of the RQ from the paper and summarizing them into a specific format, and appropriateness evaluation by machines are challenging tasks that have not yet been addressed, to our best knowledge.

RQ extraction and evaluation can be considered subtasks of document summarization, as they involve selecting and concisely expressing important information from research papers. Applying document summarization techniques may help solve these tasks with reasonable accuracy.

To improve the performance of summarization, it is generally necessary first to define a performance evaluation function and then optimize the summarization model to maximize the value of that evaluation function. For example, \citet{lewis-etal-2020-bart} used perplexity as an evaluation function to assess the similarity between human-created summaries and summaries generated by BART. In this way, identifying an appropriate evaluation function is crucial for improving the performance of RQ extraction.

An evaluation function's output must strongly correlate with human judgment to accurately measure qualitative improvements in summaries. Assessing the correlation between existing evaluation functions and human evaluation in the context of RQ is crucial. If existing functions do not correlate well, developing RQ-specific evaluation functions will be necessary.

Research on automatic evaluation of document summarization has verified the correlation between automatic evaluation functions and human evaluation \citep{fabbri2020summeval}. However, many of these studies target specific domains, such as news articles, and there may be biases specific to those domains \citep{kryscinski-etal-2020-evaluating}. Compared to news articles, research papers tend to have a more complex structure, use more technical terms, and have important information scattered throughout the document. Therefore, it is unclear how well existing automatic evaluation functions align with human judgment in RQ understanding evaluation.

Therefore, in this study, we constructed a new RQ evaluation dataset specialized for the domain of research papers. This dataset consists of:
\begin{enumerate}
    \item Paper abstracts and introductions
    \item RQ extracted by Large LLM from the abstracts and introductions
    \item Human-annotated evaluation scores for the extracted RQ, considering the abstracts and introductions
\end{enumerate}

By using this dataset, an automatic evaluation function for RQ can be established in the future, and the performance of RQ extraction can be optimized against that evaluation function. This will enable the achievement of RQ extraction, making a unique contribution that is distinct from conventional tasks dealing with research papers. While datasets for summarizing research papers exist, to our best knowledge, there is no dataset specifically designed for RQ evaluation.

In this study, we used the constructed dataset to compare the alignment of existing LLM-based evaluation functions with human judgments. Specifically, we evaluated the quality of RQ in the dataset using existing LLM-based evaluation functions and compared their evaluations with the human evaluations provided in the dataset. An overview of this evaluation procedure is illustrated in \Cref{fig:image-of-whole-process}.

The analysis revealed that LLM-based evaluation functions do not correlate with human judgment as much as previously suggested. This suggests the need to design new evaluation functions capable of handling complex tasks such as RQ understanding evaluation. Furthermore, this insight may apply to the evaluation of automatic summarization in other specialized domains, not just RQ.

The contributions of this study are as follows:
\begin{itemize}
\item We conducted human evaluations of RQ on papers in the field of machine learning and constructed a dataset containing these evaluations.
\item Using the constructed dataset, we compared the alignment of existing LLM-based evaluation functions with human judgments.
\end{itemize}

The structure of this paper is as follows. In Section 2, we discuss related work, and in Section 3, we explain the details of the proposed dataset. In Section 4, we present the experimental setup, results, and discussion. In Section 5, we provide a summary and future outlook. In Section 6, we discuss the potential risks. Finally, in Section 7, we discuss the limitations of this study.

\section{Related Work}
\label{sec:Related Work}

\subsection{Evaluation Functions}
Evaluation of automatic summarization is important for properly measuring the performance of summarization systems.
Automatic evaluation functions can be broadly divided into two types: Reference-based and Reference-free. Assuming Document is the original document, Summary is the generated summary, and Reference is the human-created summary, in the Reference-based setting, evaluation is performed using Document, Summary, and Reference. On the other hand, in the Reference-free setting, evaluation is performed using only Document and Summary \citep{sai-2022-survey-of-NLG-eval}.

In recent years, it has become clear that evaluation functions using LLM, typified by GPT-4, show higher performance than conventional evaluation functions\citep{wang-etal-2023-chatgpt,liu-etal-2023-g}, attracting much attention. LLM-based evaluation functions work by having the LLM return a score based on a prompt that includes the document to be evaluated and its summary, and in some cases, a reference summary.

LLM evaluation functions exist in both Reference-based settings and Reference-free settings. In the Reference-based setting, LLMs can more directly consider alignment with reference summaries, but evaluation needs to be performed even when reference summaries do not exist. On the other hand, in the Reference-free setting, the language understanding ability of LLMs can be utilized to directly evaluate the quality of summaries \citep{wang-etal-2023-chatgpt}.

There are various types of LLM evaluation functions, differentiated by the presence or absence of the features described in \Cref{tab:description-of-terms}. Specifically, as categorized in \Cref{tab:method-categorize}, the differences are mainly distinguished by whether they include evaluation procedures \citep{liu-etal-2023-g}, output scores in batches \citep{Yuan2023BatchEvalTH}, or require explanations for scores \citep{chiang-lee-2023-closer}. More details are provided in \Cref{sec:Evaluation Functions}.

\subsection{Datasets Targeting Academic Papers}
Building datasets targeting academic papers is one of the important research challenges in the field of natural language processing. Various datasets have been proposed, such as QASPER \citep{dasigi-etal-2021-dataset}, SciCite \citep{cohan-etal-2019-structural}, and others, each focusing on different aspects of academic papers (see \Cref{sec:appdx datasets targeting academic papers} for more details).

Unlike previous datasets, our proposed dataset manually evaluates the quality of RQ extracted by language models from paper abstracts and introductions. It quantitatively measures how closely the extracted RQ match the true RQ, directly assessing the RQ generation performance of language models. Creating this dataset involves a challenging and complex annotation process, requiring annotators to extract key information from papers, organize it, and evaluate the extracted RQ. 
\begin{table}
\centering
\small
\setlength{\tabcolsep}{3.5pt}
\begin{tabular}{p{2.5cm}>
{\arraybackslash}p{5cm}}
\hline
\textbf{Term} & \textbf{Description}\\
\hline
Document & Document (ex. paper)\\
Summary & Summary generated from the Document (ex. RQ)\\
Reference & Ground truth summary created by humans \\
Instruction & Task instructions\\
Aspect & Evaluation aspect\\
Output Space & Range of evaluation values\\
Criteria & Evaluation criteria\\
Evaluation Steps & Evaluation steps\\
Data Sample & Data unit for evaluation (ex. sample or batch)\\
Multiple Score sample & Number of samples for evaluation scores\\
Score-Explanation & Whether to have LLM explain the reasons for the evaluation scores\\
ICL & Whether in-context learning is used\\
\hline
\end{tabular}
\caption{Representative terms and their descriptions used to explain evaluation functions. As described later, each evaluation function is differentiated by the presence or absence of these elements.}
\label{tab:description-of-terms}
\end{table}

\begin{table*}
\centering
\small
\setlength{\tabcolsep}{3.5pt}

\begin{tabular}{p{3cm}>{\centering\arraybackslash}p{1.2cm}>{\centering\arraybackslash}p{1.2cm}>{\centering\arraybackslash}p{1.2cm}>{\centering\arraybackslash}p{1.2cm}>{\centering\arraybackslash}p{1.2cm}>{\centering\arraybackslash}p{1.2cm}>{\centering\arraybackslash}p{1.2cm}>{\centering\arraybackslash}p{1.2cm}>{\centering\arraybackslash}p{1.2cm}}
\hline
\textbf{Name} & \textbf{Instruction} & \textbf{Aspect} & \textbf{Output Space} & \textbf{Criteria} & \textbf{Data Sample} & \textbf{Eval-procedure} & \textbf{Multiple Score} & \textbf{Score-explain} & \textbf{ICL} \\
\hline
\citep{Liu2023CalibratingLE} & \checkmark & \checkmark & \checkmark & \dag & Single & \ding{55} & \ding{55} & \checkmark & \ding{55} \\
\citep{wang-etal-2023-chatgpt} & \checkmark & \checkmark & \checkmark & \checkmark & Single & \ding{55} & \ding{55} & \ding{55} & \ding{55} \\
\citep{liu-etal-2023-g} & \checkmark & \checkmark & \checkmark & \checkmark & Single & \checkmark & \checkmark & \ding{55} & \ding{55} \\
\citep{chiang-lee-2023-closer} & \checkmark & \checkmark & \checkmark & \checkmark & Single & \checkmark & \checkmark & \checkmark & \ding{55} \\
\citep{Yuan2023BatchEvalTH} & \checkmark & \checkmark & \checkmark & \checkmark & Batch & \ding{55} & \checkmark & \checkmark & \ding{55} \\
\citep{Gong2023CoAScoreCP} & \checkmark & \S & \checkmark & \checkmark & Single & \ding{55} & \ding{55} & \ding{55} & \ding{55} \\
\citep{jain-etal-2023-multi} & \ding{55} & \ding{55} & \ding{55} & \ding{55} & Single & \ding{55} & \ding{55} & \ding{55} & \checkmark \\
\hline
\end{tabular}
\caption{
The \checkmark in the table indicates that the element is included, while \ding{55} indicates that it is not included. Additionally, the Aspect in \citep{Gong2023CoAScoreCP} indicates estimating sub-aspects as sub-components of the Aspect, and the Criteria in \citep{Liu2023CalibratingLE} indicates using criteria estimated by GPT-4 instead of human-written descriptions.
}
\label{tab:method-categorize}
\end{table*}

\section{Proposed Dataset}

\subsection{Dataset}
In this study, we constructed a dataset targeting 104 papers accepted as long papers at ACL from 2016 to 2023. The papers subject to annotation were limited to those proposing a solution (method) to a specific problem, which is a common format for many ACL papers. In these papers, the RQ is expected to be formulated as ``Can a certain `problem' be solved by a certain `method' ? ''. 

For each paper, we used GPT-4 to estimate the RQ and collected human scores evaluating the quality of the estimated RQ.

\subsection{RQ Estimation Method}
We input the abstracts and introductions of the papers into GPT-4 and used the following three different prompts to extract three RQ for each paper. The specific prompts are listed in \Cref{sec:appdx list of prompts used to generate the RQs}. 

\begin{enumerate}
    \item prompt1: A prompt that simply instructs to estimate the RQ
    \item prompt2: A prompt that specifies the RQ should be in a specific format (``Can the PROBLEM be solved by the METHOD?'') and instructs to estimate the RQ
    \item prompt3: A prompt that applies a method to generate structured text in an XML-like markup language, while specifying that the RQ should be in a specific format
\end{enumerate}

We chose to use three different prompts in order to introduce diversity in the quality of the extracted RQ, ranging from poorly extracted to well-crafted ones. By including this variety in our dataset, we can better evaluate the performance of the LLM-based evaluation functions across a range of RQ qualities:

The third prompt is expected to seamlessly integrate Chain-of-Thought (CoT) and external tools by generating structured text in an XML-like markup language, allowing control of undesirable behaviors of language models \citep{Yamauchi2023LPMLLM}.
The characteristics of each prompt are explained in \Cref{tab:prompt-categorize}, and the actual prompts used are explained in \Cref{tab:rq-generation-prompt}. RQ generation was performed using \textit{gpt-4-0125-preview}, with a temperature of 1 and topP of 1. 

Furthermore, according to \Cref{sec:Annotation Score per prompt}, when taking the average of all annotators' annotations, prompt 3 tends to be assigned high scores for Problem Score, Method Score, and Is Target RQ Type, suggesting that it is the most effective prompt for extracting RQ.

\begin{table}
\centering
\small
\setlength{\tabcolsep}{3.5pt}
\begin{tabular}{p{1.5cm}>{\centering\arraybackslash}p{1.5cm}>
{\centering\arraybackslash}p{1.5cm}>
{\centering\arraybackslash}p{1.5cm}}
\hline
\textbf{prompt} & \textbf{Explicit RQ Type}& \textbf{Explicit RQ Nature}& \textbf{Elicit Thinking}\\
\hline
prompt1 & \ding{55}  & \checkmark & \ding{55} \\
prompt2 & \checkmark & \ding{55}  & \ding{55} \\
prompt3 & \ding{55}  & \checkmark & \checkmark\\
\hline
\end{tabular}
\caption{Categorization of prompts. Explicit RQ Type indicates that the prompt explicitly instructs the model to extracted RQ following the format "Can the PROBLEM be solved by the METHOD?". Explicit RQ Nature indicates that the prompt text itself explicitly describes the desired characteristics of the RQ to be extracted. Elicit Thinking indicates whether Chain-of-Thought (CoT) is applied during output.}
\label{tab:prompt-categorize}
\end{table}

\subsection{Annotation Method}
\subsubsection{Annotators}
The annotation was performed by a total of three people: two researchers who routinely read papers in the field of machine learning and one graduate student majoring in information science. 
All annotators were male and were not compensated for their work. The annotators were informed about the purpose of the annotation task and how the data would be used in the research. They provided verbal consent to participate in the annotation process.

\subsubsection{Annotation Perspectives}
Each annotator scored the RQ estimated by GPT-4 from the following three perspectives:
\begin{itemize}
    \item Problem Score: Does the RQ accurately estimate the true problem? (3 levels from 0 to 2)
    \item Method Score: Does the RQ accurately estimate the true method? (3 levels from 0 to 2)
    \item Is target rq type: Is the RQ in a specific format (proposing a solution to an existing problem)? (2 levels: 0 or 1)
\end{itemize}

\subsection{Annotation Results and Analysis}
In general, the annotation results for each data point can vary depending on the annotator.
Therefore, by selectively retaining data with high agreement among annotators, a highly reliable dataset can be created.

When the difficulty of annotation is relatively low and there is little variation among annotators, measures such as Krippendorff's alpha coefficient \citep{fabbri2020summeval} and perfect agreement rate \citep{ValenzuelaEscarcega2015IdentifyingMC} have been used as indicators of agreement among annotators.

On the other hand, when the difficulty of annotation is relatively high and there is greater variation among annotations, different agreement measures may be used.
For example, in the construction of the PubmedQA dataset \citep{jin-etal-2019-pubmedqa}, when the labels of Annotator 1 and Annotator 2 do not match, a discussion is held between the two to reach a consensus, and if a consensus still cannot be reached, that instance is removed.

The task in this study is a highly difficult one that requires understanding and judgment of complex text, and falls into a situation where there is large variation among annotations.

In fact, we could not find sufficient agreement among annotators using Krippendorff's alpha coefficient and perfect agreement rate.
Therefore, we decided to use the results of a majority vote as an indicator of agreement.
Specifically, we consider the annotations that two out of three annotators agree on as the ground truth (GT).
Using this method, 69.5\% of the data (217 out of 312 RQ) were adopted as GT.

\begin{table*}
\centering
\small
\setlength{\tabcolsep}{3.5pt}
\begin{tabular}{
p{4cm}>
{\arraybackslash}p{7cm}>
{\centering\arraybackslash}p{1.2cm}>
{\centering\arraybackslash}p{1.2cm}>
{\centering\arraybackslash}p{1.2cm}}
\hline
\textbf{paper-title} & \textbf{extracted RQ}& \textbf{problem score}&\textbf{method score}&\textbf{rq-format}\\
\hline
Are Training Samples Correlated? Learning to Generate Dialogue Responses with Multiple References & Can the generic response problem in open-domain dialogue generation be solved by utilizing a novel two-step generation architecture that models multiple responses jointly by considering their correlations? & 2& 2& 1\\
\hline

\hline
\end{tabular}
\caption{Example of human annotations}
\label{tab:dataset-sample}
\end{table*}

\section{Experiment}
\label{sec:Experiment}
\subsection{Evaluation Functions Used in This Experiment}
\label{sec:Evaluation Functions}
In recent years, evaluation methods using LLM have been actively researched. The automatic evaluation methods compared in the experiments are summarized in \Cref{tab:method-categorize}. 

The details of each method are provided in \Cref{sec:appdx evaluation functions}.
In the following, we explain the experimental setup and results.

\subsection{Experimental Setup}
\label{sec:Experimental Setup}
In this study, we evaluated the correlation between human-annotated scores and scores output by various LLM-based evaluation functions using \textit{gpt-4o-2024-05-13} on the dataset we created. The evaluation settings were based on previous studies. \citet{jain-etal-2023-multi} set the number of few-shots to 5, while \citet{Yuan2023BatchEvalTH} set the batch size to 10 and the number of loops to 3. Additionally, \citet{liu-etal-2023-g} and \citet{chiang-lee-2023-closer} set the output \textit{n} to 20. For other setting items such as \textit{temperature} and \textit{top\_p}, we used the values reported in each paper.

The model output obtained as a result of the evaluation may contain text as shown in \Cref{tab:example-of-output}. Therefore, it is necessary to extract the actual values from the output. This extraction process was performed using the Python code attached in \Cref{sec:pythoncode-for-parse}.

\subsection{Results}
\label{sec:main-result}

We summarized the correlation coefficients between each evaluation functions and human evaluation in Table 6. The table shows the Pearson correlation coefficients between the scores given by LLM-based evaluation functions and human evaluations for each aspect of RQ quality.

While a correlation coefficient of around 0.5 was obtained for the Method Score, the correlation coefficients for other aspects were below 0.2. This suggests that the LLM-based evaluation functions proposed so far do not correlate well with human evaluations in assessing RQ quality, particularly in aspects other than identifying the method.

In contrast, previous studies have reported that these LLM-based evaluation functions correlate to some extent with human evaluations. For example, in studies such as \citet{liu-etal-2023-g}, the correlation coefficients between automatic evaluation functions and human evaluations were around 0.6 for some aspects, and most exceeded 0.35.

Our results suggest that the correlation between previously proposed LLM-based evaluation functions and human evaluations may have been overestimated. While these evaluation functions have been reported to correlate with human judgments in tasks such as news summarization, our findings indicate that they may not yet be able to provide evaluations that correlate with human judgments for tasks beyond news summarization, such as RQ evaluation. This result implies the need to develop optimal evaluation functions for each task.

\begin{table*}
\centering
\small
\setlength{\tabcolsep}{3.5pt}

\begin{tabular}{p{3.5cm}>{\centering\arraybackslash}p{1.2cm}>{\centering\arraybackslash}p{1.2cm}>{\centering\arraybackslash}p{1.2cm}>{\centering\arraybackslash}p{1.2cm}>{\centering\arraybackslash}p{1.2cm}>
{\centering\arraybackslash}p{1.2cm}>{\centering\arraybackslash}p{1.2cm}>{\centering\arraybackslash}p{1.2cm}}
\hline
{Name} & \multicolumn{2}{c}{Problem Score} & \multicolumn{2}{c}{Method Score} & \multicolumn{2}{c}{Format Score}  \\

& $\rho$ & $\tau$ & $\rho$ & $\tau$ & $\rho$ & $\tau$  \\
\hline
\citep{Liu2023CalibratingLE} & 0.120 & 0.114 & nan& nan & -0.031 & -0.031  \\
\citep{wang-etal-2023-chatgpt} & 0.076 & 0.071 & 0.0176& 0.0165 & 0.0248 & 0.0236  \\
\citep{chiang-lee-2023-closer} $\alpha$ & 0.121 & 0.091 & \textbf{0.493}& \textbf{0.405} & 0.067 & 0.055  \\
\citep{chiang-lee-2023-closer} $\beta$ & 0.110 & 0.088 & 0.281& 0.233 & \textbf{0.139} & 0.125  \\
\citep{liu-etal-2023-g} & \textbf{0.214} & \textbf{0.167} & 0.227& 0.185 & 0.121 & 0.108  \\
\citep{Yuan2023BatchEvalTH} & 0.041 & 0.039 & 0.149& 0.143 & 0.006 & 0.005  \\
\citep{Gong2023CoAScoreCP} & -0.048 & -0.045 & 0.165& 0.160 & -0.134 & -0.134  \\
\citep{jain-etal-2023-multi} & -0.096 & -0.086 & 0.101& 0.094 & 0.126 & \textbf{0.126}  \\
\hline
\end{tabular}
\caption{A list of correlation coefficients. Following \citet{liu-etal-2023-g}, we calculated the Spearman and Kendall-Tau correlation coefficients. $\rho$ denotes the Spearman correlation coefficient, and $\tau$ denotes the Kendall-Tau correlation coefficient. Additionally, $\alpha$ refers to analyze-rate from \citet{liu-etal-2023-g}, and $\beta$ refers to rate-explain. In analyze-rate, the LLM first analyzes the input information, points out the good and bad points, and then outputs the final evaluation score. On the other hand, in rate-explain, the LLM outputs the evaluation score based on the input information first, and then explains the rationale for the evaluation. For the Method Score from \citep{Liu2023CalibratingLE}, since only the same value was output, the correlation coefficient is nan.}
\label{tab:result}
\end{table*}

\subsection{Discussion}
This section investigates common tendencies across evaluation methods, examines method reproducibility, and analyzes performance improvement strategies.

We first analyze similarities in incorrectly evaluated RQ sets for each method and the impact of input/output token count on performance.

Next, we discuss result variability due to sample size and model differences when assessing method reproducibility.

Finally, we confirm the importance of modeling evaluation procedures and verify how increasing procedure steps affects performance. We also fine-tune models to test the hypothesis that directly learning scoring patterns from data outperforms prompt-based methods.

\subsubsection{Investigating Common Tendencies across Evaluation Methods}
In this section, we analyze the common tendencies across evaluation methods from two perspectives: examining the similarity of RQ sets with incorrect evaluation values and investigating the impact of input and output token counts on performance. These analyses aim to clarify common tendencies and provide insights for improving future evaluation methods.

\begin{figure}[t]
\includegraphics[width=\linewidth]{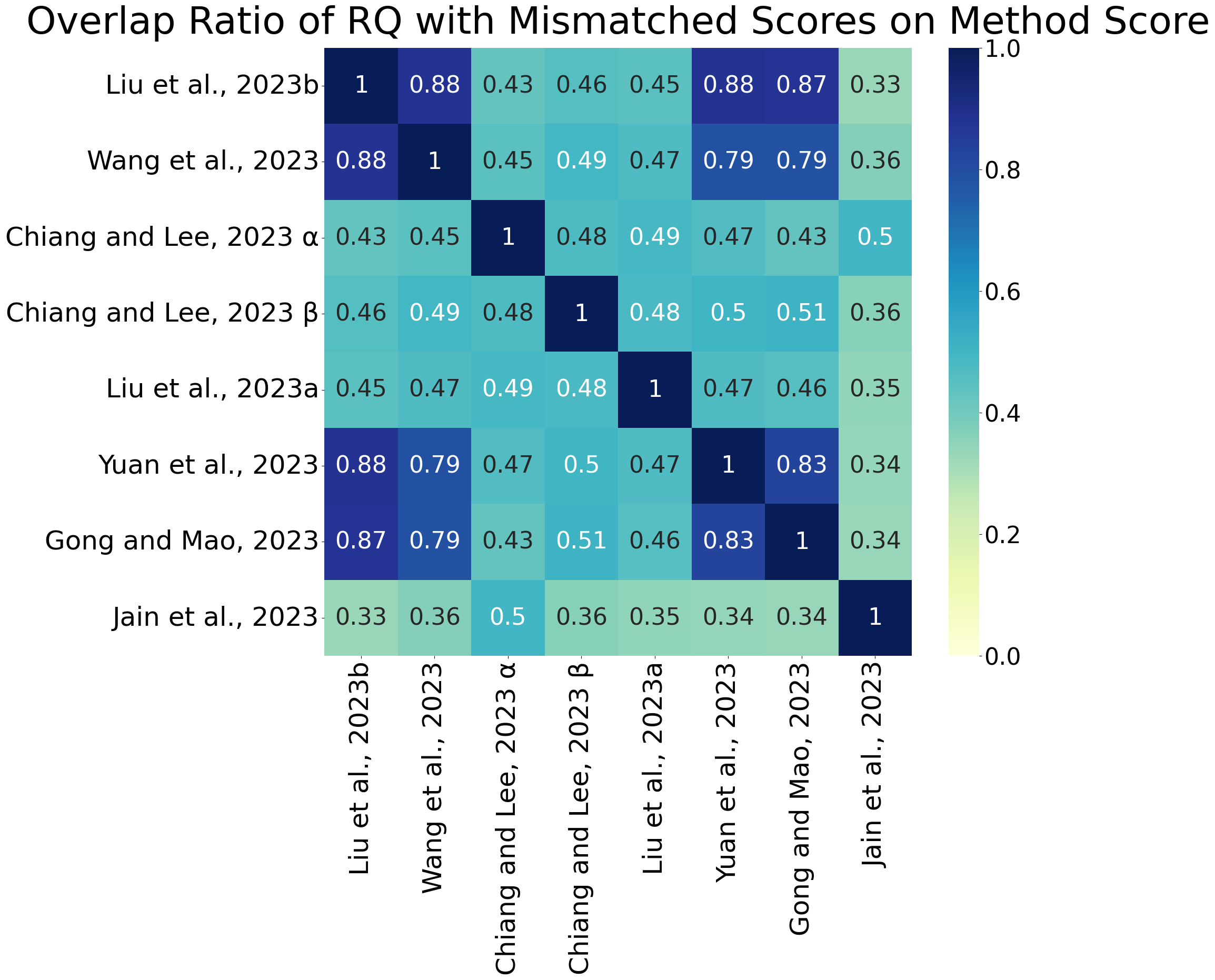}
\caption {Visualization of the overlap rate of RQ with mismatched evaluation values between methods, categorized by Method Score, as a correlation diagram.}
\label{fig:overlap-of-miss}
\end{figure}

\paragraph{How similar are the sets of RQ for which incorrect evaluation values were outputted between methods?}
\subparagraph{Analysis of Common Patterns in Misclassified RQ}
We hypothesized that there might be a trend in the RQ with errors, where errors are defined as estimated scores different from the GT. \Cref{fig:overlap-of-miss} visualizes the overlap rate between RQ sets with mismatched evaluation values, categorized by score type. To account for varying output ranges, we set thresholds using percentiles and converted them into three or two values. For Problem Score and Method Score, the overlap rate of RQ with errors was high in \citet{liu-etal-2023-g}, \citet{wang-etal-2023-chatgpt}, \citet{Yuan2023BatchEvalTH}, and \citet{Gong2023CoAScoreCP}, suggesting common issues leading to similar errors. The analysis for Problem Score and Is target rq type is in the appendix.

\subparagraph{Correlation Analysis with Paper Acceptance Year and Length}
We analyzed the characteristics of commonly misclassified RQ, hypothesizing that the year of paper acceptance and paper length might be related. However, the analysis showed no clear trends, indicating that these factors do not explain the characteristics of RQ with errors.

\paragraph{Do more tokens lead to better performance?}
The previous analysis did not identify any factors that could adequately explain the characteristics of frequently misclassified RQ. Consequently, based on the performance difference between \citet{chiang-lee-2023-closer} and \citet{Yuan2023BatchEvalTH}, we hypothesized that input and output token counts influence model performance. However, visualizing the relationship between token counts and manual evaluation revealed no clear correlation. Details are provided in \Cref{sec:Do more tokens lead to better performance?}.

\subsubsection{Reproducibility of the Methods}
\begin{figure}[t]
\includegraphics[width=\linewidth]{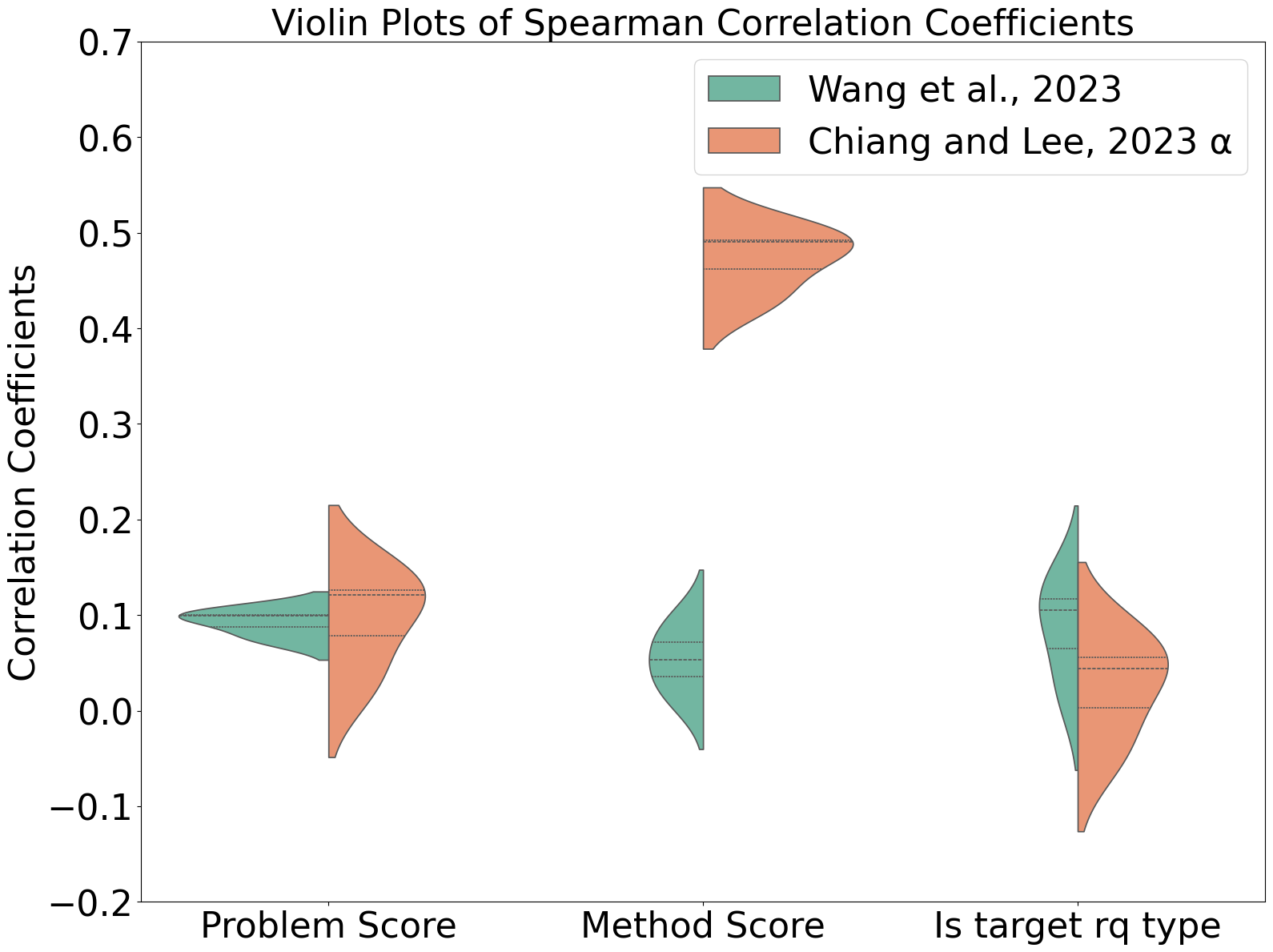}\hfill
\caption {Visualization of Spearman correlation coefficients using violin plots, comparing \citet{wang-etal-2023-chatgpt} and \citet{chiang-lee-2023-closer} to confirm the variability due to differences in sample count when the temperature is set to 1 for both methods. Visualization of Kendall correlation coefficients is shown in \Cref{sec:appdx Impact of sample count on result variability}.}
\label{fig:violinplot_diff_n_sample}
\end{figure}

\begin{figure}[t]
\includegraphics[width=\linewidth]{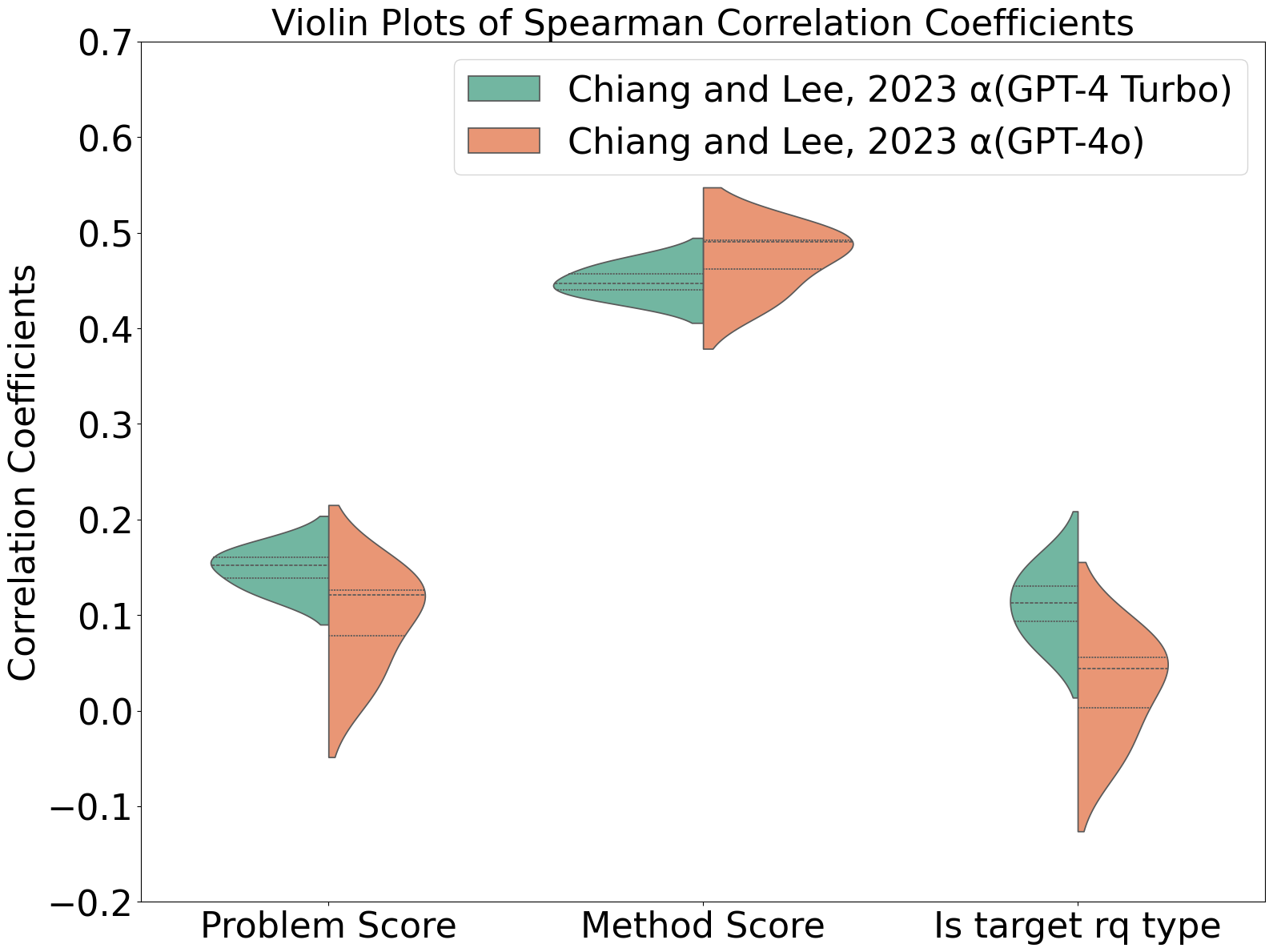}\hfill
\caption {Visualization of Spearman correlation coefficients using violin plots, comparing the analyze-rate of \citet{chiang-lee-2023-closer} to confirm the variability due to differences between \textit{gpt-4-turbo-2024-04-09} and \textit{gpt-4o-2024-05-13} when the temperature is set to 1 for both. Visualization of Kendall correlation coefficients is shown in \Cref{sec:appdx Variability of results due to model differences}.}
\label{fig:violinplot_diff_model}
\end{figure}

In this study, we define reproducibility as the ability to obtain consistent results when repeating an experiment under the same conditions. To the best of our knowledge, reproducibility has not been sufficiently discussed in the context of LLM-based evaluation functions for text generation, despite its importance. We investigate the reproducibility of the methods from two perspectives: the impact of sample count on result variability and the variability due to model differences.
\paragraph{Impact of sample count on result variability}
The number of samples from LLM outputs may differ depending on the evaluation function (\Cref{tab:method-categorize}). We hypothesized that methods with multiple sampling would have less variability in the results. We conducted three trials each for \citet{wang-etal-2023-chatgpt} (single sampling) and \citet{chiang-lee-2023-closer} (20 samplings) the same temperature of 1 and visualized the variability of Spearman correlation coefficients using violin plots. Contrary to our hypothesis, \citet{chiang-lee-2023-closer} sometimes showed more variability (\Cref{fig:violinplot_diff_n_sample}), suggesting that multiple sampling may not significantly improve result reliability.
\paragraph{Variability of results due to model differences}
We used \textit{gpt-4o-2024-05-13} in this study (\Cref{sec:Experimental Setup}), but it is necessary to confirm the extent to which evaluation values change when using a different model. We hypothesized that the reliability of results may vary depending on the model, even with the same method. We compared the variability of Spearman correlation coefficients between \textit{gpt-4-turbo-2024-04-09} and \textit{gpt-4o-2024-05-13} for the analyze-rate of \citet{chiang-lee-2023-closer} (best-performing method) with a temperature of 1. While there was no significant difference in performance, \textit{gpt-4-turbo-2024-04-09} showed less variability (\Cref{fig:violinplot_diff_model}), suggesting that output results may fluctuate even if model performance does not vary significantly.

\subsubsection{Analysis for Performance Improvement}
This section describes our attempts to improve performance. We qualitatively analyzed the best-performing methods, confirming the importance of modeling the evaluation procedure. To improve the completeness of the evaluation procedure, we increased the number of steps and verified the impact on performance. Finally, we attempted fine-tuning to test the hypothesis that directly learning scoring trends from data leads to higher performance than prompting-based methods.
\paragraph{Importance of modeling the evaluation procedure}
The best-performing methods, \citet{chiang-lee-2023-closer} and \citet{liu-etal-2023-g}, estimate the evaluation procedure in a two-step process. First, they estimate the procedure itself, then calculate the final evaluation value based on the estimated procedure. Qualitatively, this method largely reproduces the annotation process (see \Cref{sec:appdx sample evaluation procedure}), suggesting that reproducing the annotation process through modeling may be important for this task.
\paragraph{Impact of Increasing the Number of Evaluation Procedure Steps on Performance}

While modeling the evaluation procedure positively impacted performance, some steps were omitted compared to our actual annotation guidelines, as discussed in \Cref{sec:appdx sample evaluation procedure}. We hypothesized that increasing the number of steps in the estimated procedure could capture these omitted processes, potentially improving performance by more closely mimicking our evaluation process. We studied the extent to which performance improves by increasing the number of steps from the original methods (yielding about 5 steps, see \Cref{sec:appdx sample evaluation procedure}) to 10 and 30 steps.

However, despite the increased number of steps, the LLM was unable to fully reproduce the omitted evaluation procedures as shown in\Cref{sec:Impact of Increasing the Number of Evaluation Procedure Steps on Performance}. This suggests that the LLM lacks the specialized knowledge of how researchers read and analyze papers, which cannot be compensated for by arbitrarily increasing the number of steps in the evaluation procedure. 

\paragraph{Correcting the Overestimation of RQ Scores}
In \Cref{sec:appdx overestimation of score}, we analyze the difference between the estimated values and the GT for each method. The results show that the estimated values of \citet{chiang-lee-2023-closer} and \citet{liu-etal-2023-g} may be overestimated compared to other methods. In other words, if we can suppress this overestimation by some method, the correlation may improve. In the future, we need to explore methods to reduce this overestimation.

\paragraph{Learning the scoring patterns from the dataset}
While the evaluation functions used in this experiment attempt to improve performance through prompting using GPT-4, we hypothesized that directly learning the scoring trends from the dataset would lead to better results. 

To test this, we fine-tuned the open-source LLM Gemma-7b using LoRA, but the results were not promising (see \Cref{sec:appdx FT on Gemma} for details). In future research, we plan to explore this approach using larger models.

\section{Conclusion}
In this study, we constructed a new dataset that pairs RQ extracted by GPT-4 with their manual evaluations, targeting papers accepted at ACL. Using this dataset, we studied the correlation of GPT-4-based automatic evaluation functions with human evaluation.

Our experiments revealed that the automatic evaluation functions, which were reported to have high correlation with human annotators in existing studies, showed only low correlation in the RQ evaluation task. This suggests the possibility that appropriate evaluation functions differ depending on the task, supporting the significance of creating and publishing a dataset with human annotations. On the other hand, the method that estimates the evaluation procedure showed relatively high performance in evaluating the Method Score of RQ.

The results of this study provide insights for the development of automatic evaluation functions in the RQ generation task for papers. In the future, the design of evaluation functions specialized for the paper domain and the identification of factors contributing to the performance improvement of evaluation functions are expected.

\section{Potential risks}
Our approach uses LLM, which may disadvantage organizations that can't afford them. To address this, we should make these methods widely accessible and explore non-LLM alternatives. Additionally, Over-reliance on automatic RQ extraction might weaken researchers' skills. Therefore, researchers should use these tools to complement their expertise, ensuring they continue to develop their own capabilities.

\section{Limitations}
\subsection{Limitations of the Dataset}
The dataset constructed in this study is limited to 104 papers in the field of machine learning. Including papers from fields other than machine learning could lead to the development of models that can be commonly used across various fields, not limited to machine learning. However, due to resource constraints, we were unable to carry out such an expansion in this study. In the future, there is a need to construct datasets targeting a wider range of research fields.

Furthermore, regarding annotation, there is a possibility that it was difficult to achieve alignment among annotators because there is no firm definition of RQ and their components in the field of machine learning. The definition of RQ may vary from paper to paper, and their components encompass a wide range of aspects, leading to the possibility of different interpretations among annotators. In addition, understanding papers requires specialized knowledge, so differences in the background knowledge of annotators may have influenced the evaluation. In the future, research is needed to organize RQ and their components, particularly in the field of machine learning.

\subsection{Limitations of Evaluation}
In this study, we only conducted evaluations using GPT-4 and were unable to perform evaluations using other LLMs. Conducting evaluations using LLMs other than GPT-4 may provide deeper insights into the performance and characteristics of evaluation functions. In the future, evaluations using a variety of LLMs will be required.

Moreover, this study was limited to testing LLM-based evaluation functions developed in domains such as news article summarization, and we were unable to propose new evaluation functions that surpass their performance. These existing evaluation functions may not be suitable for evaluating complex targets like RQ in papers. RQ are composed of various components, and understanding the relationships and context between these components is required. Additionally, understanding the technical terms of papers is necessary. Therefore, in the future, it is necessary to develop evaluation functions specialized for RQ evaluation in papers, utilizing the insights obtained in this study.

\bibliography{main}

\appendix
\section{Appendix}
\subsection{Related Work}
\subsubsection{Datasets Targeting Academic Papers}
\label{sec:appdx datasets targeting academic papers}
Previously proposed datasets include QASPER \citep{dasigi-etal-2021-dataset}, SciCite \citep{cohan-etal-2019-structural}, Meaningful Citations Data Set \citep{ValenzuelaEscarcega2015IdentifyingMC}, PubMedQA \citep{jin-etal-2019-pubmedqa}, PeerRead \citep{kang-etal-2018-dataset}, and SciFact \citep{wadden-etal-2020-fact}.
QASPER is a QA dataset targeting papers, consisting of questions created by NLP experts who read only the titles and summaries of papers, and answers and supporting evidence provided by other NLP experts who read the entire papers.

SciCite is a dataset that pairs citation sentences in scientific papers with labels of their citation intent (background information, use of methods, comparison of results, etc.) and can be used for tasks such as classifying citation sentences and predicting citation intent. The Meaningful Citations Data Set is a dataset with labels identifying important citations in academic literature.

PubMedQA consists of answers from three values ("yes/no/maybe") to questions created from the titles and abstracts of medical papers. This dataset can be used to develop content understanding and question-answering systems for medical papers.

PeerRead contains 14,700 papers submitted to top conferences (ACL, NeurIPS, ICLR), their acceptance/rejection results, and peer review results by 10,700 experts. This dataset is expected to be applied to tasks such as automatic paper evaluation and peer review automation.

SciFact consists of 1,400 annotated abstracts with scientific claims and supporting evidence, with each abstract labeled as supporting or refuting the claim. This dataset can be used for tasks such as determining the veracity of claims and automatically extracting evidence.

\subsection{Dataset Creation}
\subsubsection{Data Selection Criteria}
In this study, we constructed a dataset consisting of 104 long papers accepted at ACL from 2016 to 2023.
We focused on papers published from 2016 onwards because these papers are licensed under the Creative Commons Attribution 4.0 International License. This license permits the modification of published works, which is essential for our annotation process. By selecting papers from this period, we ensure that our dataset construction and annotation efforts comply with the legal permissions granted by the license.
\subsubsection{Explanation of Dataset Rights}
The publicly released dataset includes appropriate citation information for the research papers. Additionally, this dataset targets papers published under the Creative Commons Attribution 4.0 license, and have been modified. Consequently, the dataset we have created is also subject to the Creative Commons Attribution 4.0 license.

\subsubsection{List of prompts used to extract the RQ}
\label{sec:appdx list of prompts used to generate the RQs}
As shown in \Cref{tab:rq-generation-prompt}. Prompt 3, which could not fit in this table, is mentioned in \Cref{sec:example of prompt3}.
\begin{table*}
\centering
\small
\setlength{\tabcolsep}{3.5pt}
\begin{tabular}{
p{2.5cm}>
{\arraybackslash}p{12cm}}
\hline
\textbf{prompt} & \textbf{content}\\
\hline
prompt1 & You are a friendly and helpful research assistant, specifically tasked with analyzing academic papers on machine learning, provided by users. Your sole responsibility is to meticulously read the abstracts and introductions of these papers and, using logical reasoning, deduce exactly four key research questions from each paper. It is crucial that these research questions be precise inquiries, capable of yielding empirical answers and often illuminating novel challenges that may have been previously overlooked in existing research. Your output should consist exclusively of these one research questions per paper, without any additional information or analysis. \\
\hline
prompt2 & You are a friendly and helpful research assistant, specifically tasked with analyzing academic papers on machine learning, provided by users. Your sole responsibility is to read the abstracts and introductions of these papers and, deduce exactly one key research questions from each paper. The research question has the following format. "Can the PROBLEM be solved by the METHOD?".  \\
\hline
prompt3 & \Cref{sec:example of prompt3}  \\
\hline
\end{tabular}
\caption{List of prompts used to extract the RQ in this paper.}
\label{tab:rq-generation-prompt}
\end{table*}

\subsubsection{Example of prompt3}
\label{sec:example of prompt3}
Prompt 3, which could not fit in \Cref{tab:rq-generation-prompt}, is as follows:
\begin{lstlisting}[language=Python]
<RULE>
The system and the assistant exchange messages.
All messages MUST be formatted in XML format. XML element ::= <tag attribute="value">content</tag>
Tags determine the meaning and function of the content. The content must not contradict the definition of the tag.
</RULE>

<TAG name="RULE">
This tag defines rules. The defined content is absolute.
Attributes:
    - role (optional) : A role that should follow the rules. Roles are "system" or "assistant".
Notes:
    - The assistant must not use this tag.
</TAG>

<TAG name="TAG">
This tag defines a tag. The defined content is absolute.
Attributes:
    - name : A tag name.
Notes:
    - The assistant must not use this tag.
</TAG>

<TAG name="SYSTEM">
This tag represents a system message.
Notes:
    - The assistant MUST NOT use this tag.
</TAG>

<TAG name="EOS">
Indicates the end of a message.
</TAG>

<TAG name="THINK">
This tag represents a thought process.
If you use this tag, take a drop deep breath and work on the problem step-by-step.
Attributes:
    - label (optional) : A label summarizing the contents.
Notes:
    - The thought process must be described step by step.
    - Premises in reasoning must be made as explicit as possible. That is, there should be no leaps of reasoning.
</TAG>

<TAG name="PROBLEM">
This tag represents the problem being attempted to be solved in the paper.
</TAG>

<TAG name="METHOD">
This tag represents the method or hypothesis used by the authors of the paper to solve PROBLEM.
</TAG>

<TAG name="RESEARCH_QUESTION">
This tag represents a resaerch question.
A research question is a combination of a problem to be solved and a hypothesis or method to approach it.
The general form of a research question is as follows.
- Can the PROBLEM be solved by the METHOD?
- Can the PROBLEM be explained by the METHOD?
Notes:
    - This tag must contain one PROBLEM and one METHOD tag inside.
    - The assistant must then combine the contents of the PROBLEM and METHOD and present the research question as a concise statement.
</TAG>

<RULE role="assistant">
The assistant is a friendly and helpful research assistant, specifically tasked with analyzing academic papers on machine learning, provided by users.
The assistants sole responsibility is to meticulously read the abstracts and introductions of these papers and, using logical reasoning, deduce exactly a key research questions from the paper.
The assistant first carefully reads the paper using the THINK tag, and then extracts the research questions in the paper by using the RESEARCH_QUESTION tag.
Note:
    - The assistant MUST use THINK tags before using RESEARCH_QUESTION tag.
    - The assistant MUST analyze the paper from different perspectives and extract ONE research questions.
    - It is crucial that these research questions be precise inquiries, capable of yielding empirical answers and often illuminating novel challenges that may have been previously overlooked in existing research.
    - The assistant should output only the information that can be read from PAPER; no additional information or analysis is needed.
</RULE>


\end{lstlisting}

\subsubsection{Annotation Score per prompt}
\label{sec:Annotation Score per prompt}
\begin{figure*}[t]
\includegraphics[width=0.48\linewidth]{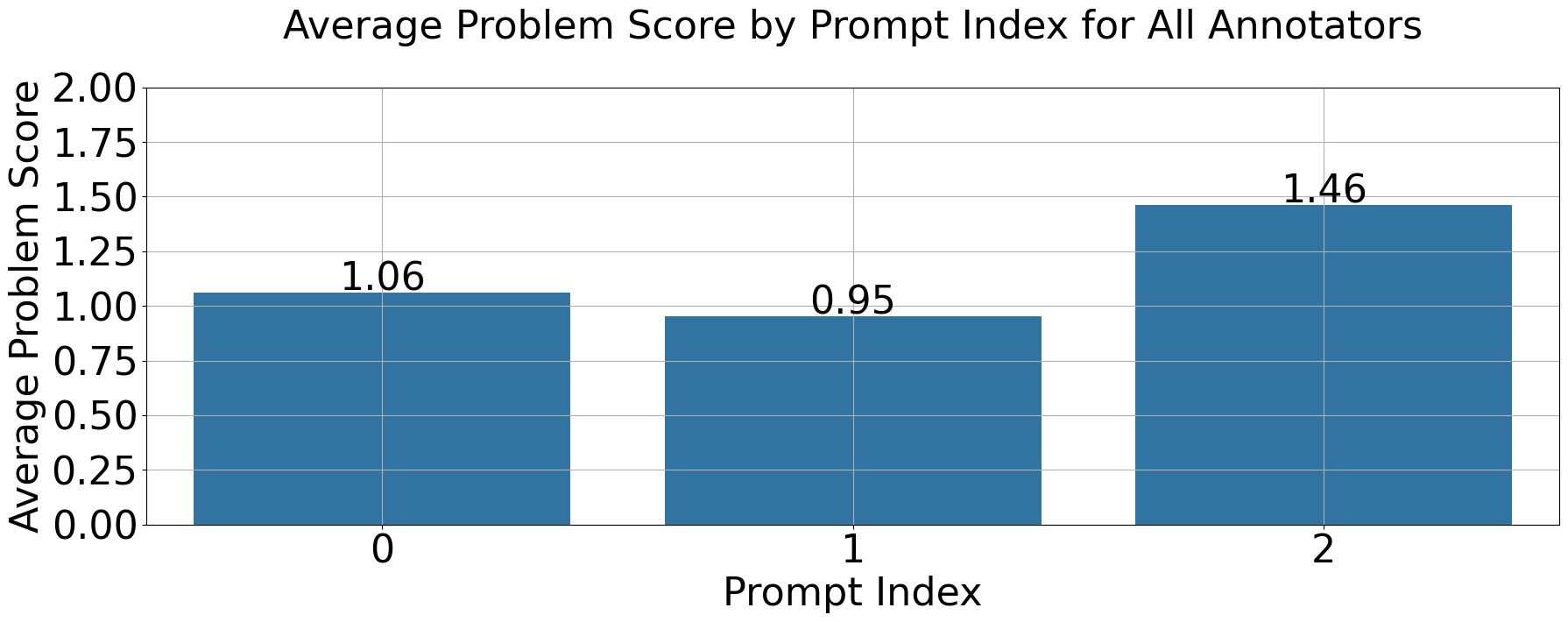}
\includegraphics[width=0.48\linewidth]{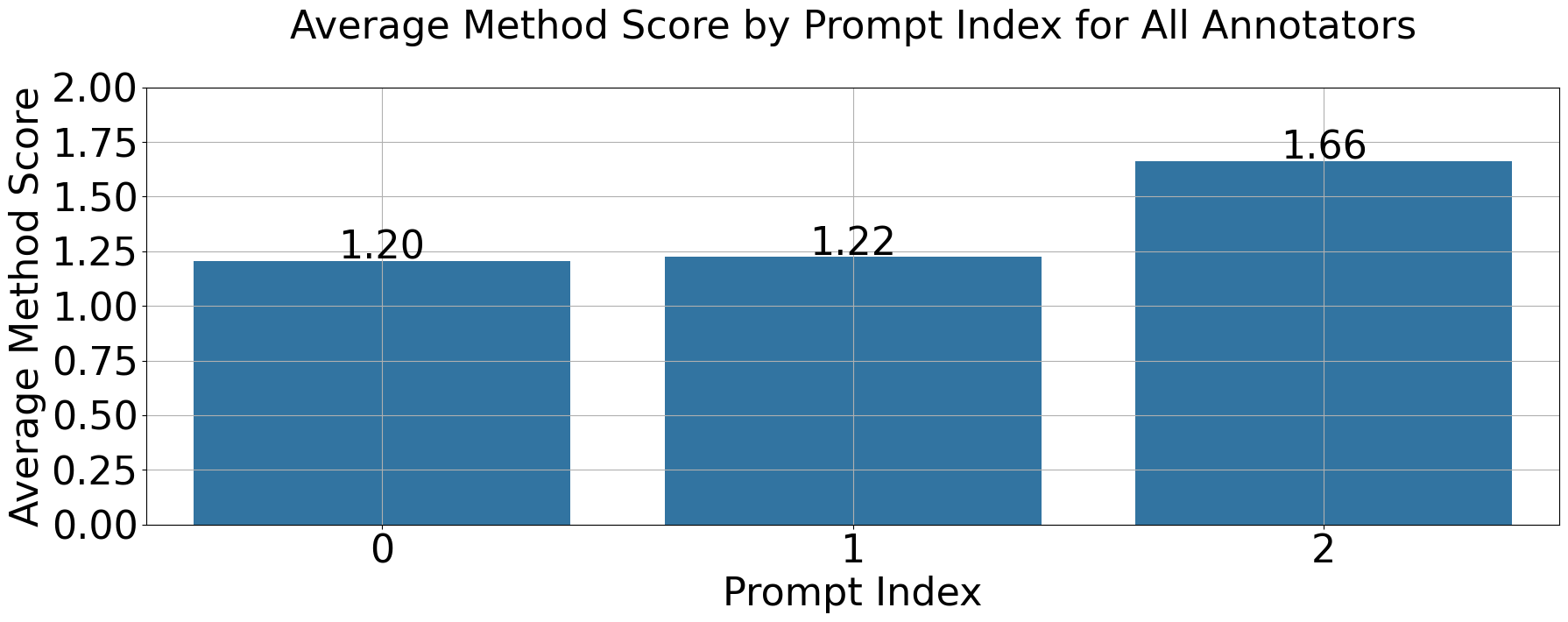}
\includegraphics[width=0.48\linewidth]{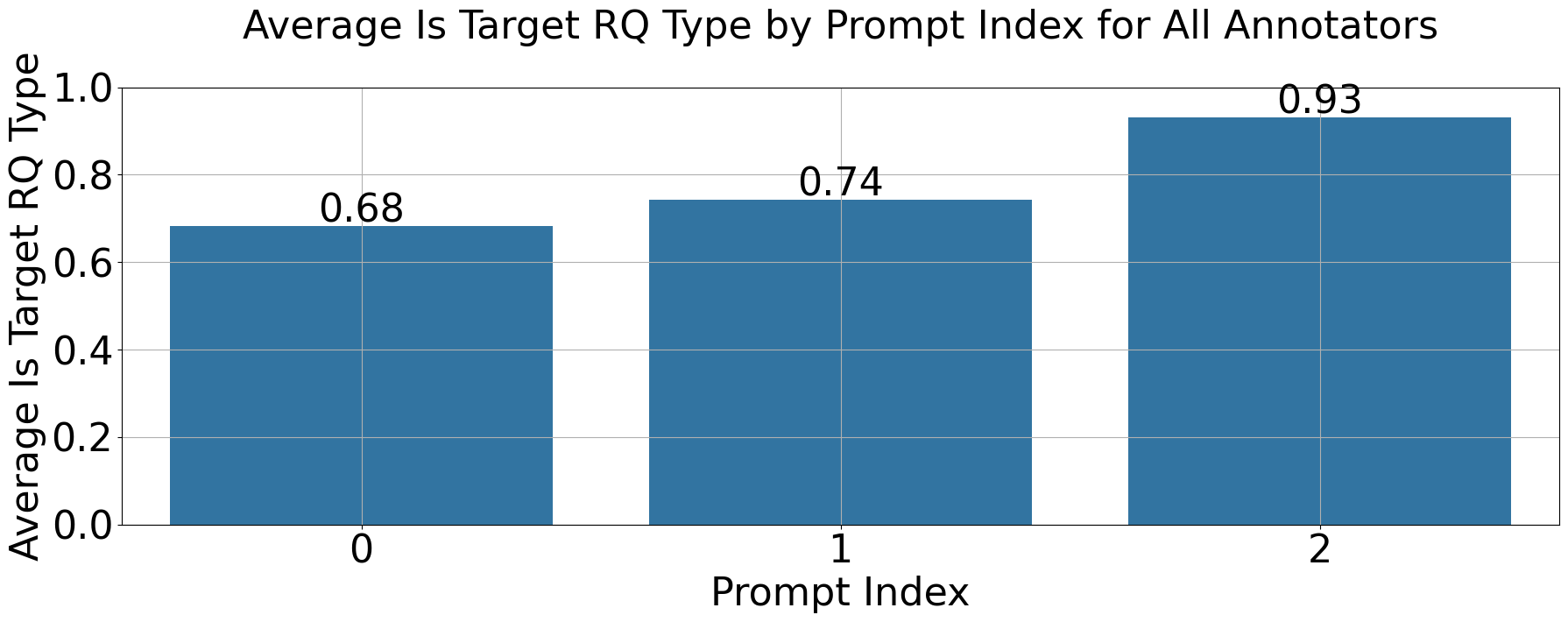}
\caption {Graph visualizing the average scores of all annotators for each prompt used to extract RQ, categorized by Problem Score, Method Score, and Is Target RQ Type.}
\label{fig:anno_score_per_prompt}
\end{figure*}

The annotation scores for each prompt used to extract RQ are visualized by the average values of all annotators for Problem Score, Method Score, and Is Target RQ Type. According to \Cref{fig:anno_score_per_prompt}, the values for prompt 3 are relatively better overall, indicating that prompt 3 has the best performance as a prompt for extracting RQ.

\subsection{Evaluation}
\subsubsection{Evaluation Functions Used in This Experiment}
\label{sec:appdx evaluation functions}
\citet{Liu2023CalibratingLE} propose a method called AUTOCALIBRATE. In this method, an arbitrary dataset labeled by human experts is first divided into training data and evaluation data. Next, the training data is used to have the LLM create its own scoring criteria. After that, the criteria are narrowed down and refined to create an evaluator closer to human judgment.

\citet{wang-etal-2023-chatgpt} propose a method to evaluate based on LLMs using human-created Aspects and Criteria. They conduct experiments in both reference-based and reference-free settings.

\citet{liu-etal-2023-g} propose a method called G-Eval. This method is characterized by having LLMs create evaluation procedures based on human-created Aspects and Criteria, and then evaluate using those evaluation procedures.

\citet{chiang-lee-2023-closer}, like \citet{liu-etal-2023-g}, have LLMs create evaluation procedures based on human-created Aspects and Criteria. However, this method is characterized by requiring explanations for the evaluations. They apply two settings: one where the evaluation explanation is analyzed before outputting the evaluation value, and another where the evaluation value is output first and then the evaluation explanation is provided.

\citet{Yuan2023BatchEvalTH} propose a method called BatchEval. This method is characterized by evaluating in batch units, taking multiple Document and Summary pairs as input.

\citet{Gong2023CoAScoreCP} propose a method called CoAScore. This method assumes that Aspects have multiple sub-aspects as lower-level perspectives, and evaluates aspects based on the evaluation values for each inferred sub-aspect.

\citet{jain-etal-2023-multi} propose a method that teaches LLMs evaluation tendencies through few-shot learning. In this case, we performed few-shot learning using a set of Document, Summary, and human annotation as one unit. This method is characterized by not using human-created Aspects or Criteria.
As described above, there are various methods to evaluation methods using LLM. These methods have their own characteristics, such as aiming for evaluations closer to human judgment, requiring explanations for evaluations, and evaluating from multiple perspectives.

\subsubsection{Example output from evaluation function}
\label{sec:appdx sample output}
The example output when using \citet{wang-etal-2023-chatgpt} is shown in \Cref{tab:example-of-output}. From this output, it is necessary to obtain the actual evaluation values using \Cref{sec:pythoncode-for-parse}.

\begin{table*}
\centering
\small
\setlength{\tabcolsep}{3.5pt}
\begin{tabular}{
p{2.5cm}>
{\arraybackslash}p{12cm}}
\hline
\textbf{Name} & \textbf{output}\\
\hline
\citep{wang-etal-2023-chatgpt} & Given the Document and the Research Question (RQ), the score for Appropriateness of Problem Description in the Summary (RQ) would be:**95****Reasoning:**The Research Question (RQ) asks how effectively a data augmentation method involving randomly switching entities can mitigate biases related to gender and sentiment in the translation of sentences containing person names without compromising translation quality. The Document comprehensively describes the following key issues:1. NMT systems exhibit gender and sentiment biases, particularly in the translation of person names.2. Gender bias is highlighted, with a specific focus on the incorrect estimation of the gender of names by NMT systems, especially for female names.3. Sentiment bias is also discussed, noting that the sentiment of translations can be affected by the presence of person names.4. A data augmentation method (switch-entity) is proposed to mitigate these biases.5. The effectiveness of this method is emphasized, along with the fact that it does not compromise translation quality.The RQ succinctly captures these aspects: it specifies the use of a data augmentation method to address gender and sentiment biases in NMT systems, and it mentions the concern about maintaining translation quality. The Document provides detailed context and evidence for these points, making the RQ highly appropriate.The reason for not giving a perfect score (100) is that while the RQ is highly aligned with the Document, it could be slightly more explicit about the specifics of the biases and the exact mechanism of the proposed data augmentation method described in the Document. However, this is a minor point, and overall the appropriateness of the problem description in the summary is excellent. \\

\hline
\end{tabular}
\caption{Example of model output in evaluation. As evaluate values are embedded within the output sentences, it is necessary to extract them.}
\label{tab:example-of-output}
\end{table*}

\subsubsection{Criteria manually created for this experiment}
\label{sec:appdx sample criteria}
As shown in \Cref{tab:criteria}.

\begin{table*}
\centering
\small
\setlength{\tabcolsep}{3.5pt}
\begin{tabular}{
p{2.5cm}>
{\arraybackslash}p{12cm}}
\hline
\textbf{Name} & \textbf{output}\\
\hline
Score Problem & Evaluates the extent to which the problem being attempted to be solved in the paper is accurately and sufficiently described in the RQ extracted by the LLM.\\
\hline
Score Method & Evaluates the extent to which the method or hypothesis used by the authors of the paper to solve the problem is accurately and sufficiently described in the RQ extracted by the LLM.\\
\hline
Is target rq type & Evaluates whether the RQ extracted by the LLM conveys the same meaning and intent as the format 'Can the PROBLEM be solved/explained by the METHOD?', without strictly adhering to this exact phrasing. The RQ does not necessarily need to follow this format word-for-word as long as it expresses the same overall idea. The accuracy of the content itself is not considered in this aspect.\\

\hline
\end{tabular}
\caption{Criteria created by humans}
\label{tab:criteria}
\end{table*}

\subsubsection{Estimated Evaluation Procedure}
\label{sec:appdx sample evaluation procedure}
The estimated evaluation procedure used in \citet{chiang-lee-2023-closer} and \citet{liu-etal-2023-g} is as shown in \Cref{tab:example-of-evaluation-step}.

The estimated evaluation procedure (see \Cref{tab:example-of-evaluation-step}) is similar to our evaluation procedure (see \Cref{paragraph : Evaluation Procedure of Problem Score}) in that it identifies the problem from the abstract and introduction of the paper, understands the scope of that problem, and then evaluates whether the RQ in the summary appropriately reflects that problem. Both approaches focus on the specific problem addressed in the paper and assess how accurately the RQ captures that problem.

However, the estimated evaluation procedure does not include an explicit step to confirm that the targeted problem is correct, as our evaluation procedure does. By adding this confirmation step, the estimated evaluation procedure could be made more complete.

\begin{table*}
\centering
\small
\setlength{\tabcolsep}{3.5pt}
\begin{tabular}{
p{0.1cm}>
{\arraybackslash}p{15cm}}
\hline
& \textbf{inferred evaluation step}\\
\hline
 & To evaluate the Appropriateness of Problem Description in the Summary (RQ), follow these steps:
 
 1. **Identify the Problem in the Document/Source Text:** Carefully read the abstract and introduction to pinpoint the central problem or issue that the paper aims to address.
 
 2. **Understand the Scope of the Problem:** Determine the extent, context, and relevance of the problem as described in the paper. Pay attention to whether the problem is well-defined and specific.
 
 3. **Compare with the RQ:** Examine the RQ to see if it accurately reflects the problem described in the Document/Source Text. Check if the RQ captures the essence and scope of the problem.
 
 4. **Assess Completeness:** Evaluate whether the RQ includes all critical aspects of the problem. Consider if any key elements or details of the problem are missing or misrepresented.
 
 5. **Rate the Appropriateness:** Based on the comparison, rate the RQ on a scale of 1 to 5 for the Appropriateness of Problem Description:
 
 - **1:** The RQ poorly describes the problem or is completely inaccurate.
 
 - **2:** The RQ describes the problem but misses several key aspects.
 
 - **3:** The RQ captures the general essence of the problem but lacks some important details.
 
 - **4:** The RQ accurately describes the problem with minor omissions or misinterpretations.
 
 - **5:** The RQ perfectly and comprehensively describes the problem as presented in the Document/Source Text.
 By following these steps, you can systematically evaluate how well the RQ captures the problem described in the paper. \\

\hline
\end{tabular}
\caption{An example of the evaluation procedure used to calculate the problem score in \citet{liu-etal-2023-g} and \citet{chiang-lee-2023-closer}}
\label{tab:example-of-evaluation-step}
\end{table*}

\subsubsection{Python snippet for parsing evaluation values from the output results}
\label{sec:pythoncode-for-parse}
\begin{lstlisting}[language=Python]
def get_score(text):
    patterns = [
        r"Score:\s*(?:.*?:\s*)?(\d+)",
        r"Score \(1-5\):\s*(\d+)",
        r"Rating:\s*(?:.*?:\s*)?(\d+)",
        r"\b(\d+)\b",
        r":\s*(\d+)"
    ]

    for pattern in patterns:
        match = re.search(pattern, text)
        if match:
            rating = match.group(1)
            return float(rating)
    
    return np.nan
\end{lstlisting}

\subsection{Discussion}
\subsubsection{Analysis of Common Patterns in Misclassified RQ for Is target rq type}
\begin{figure*}[t]
\includegraphics[width=0.48\linewidth]{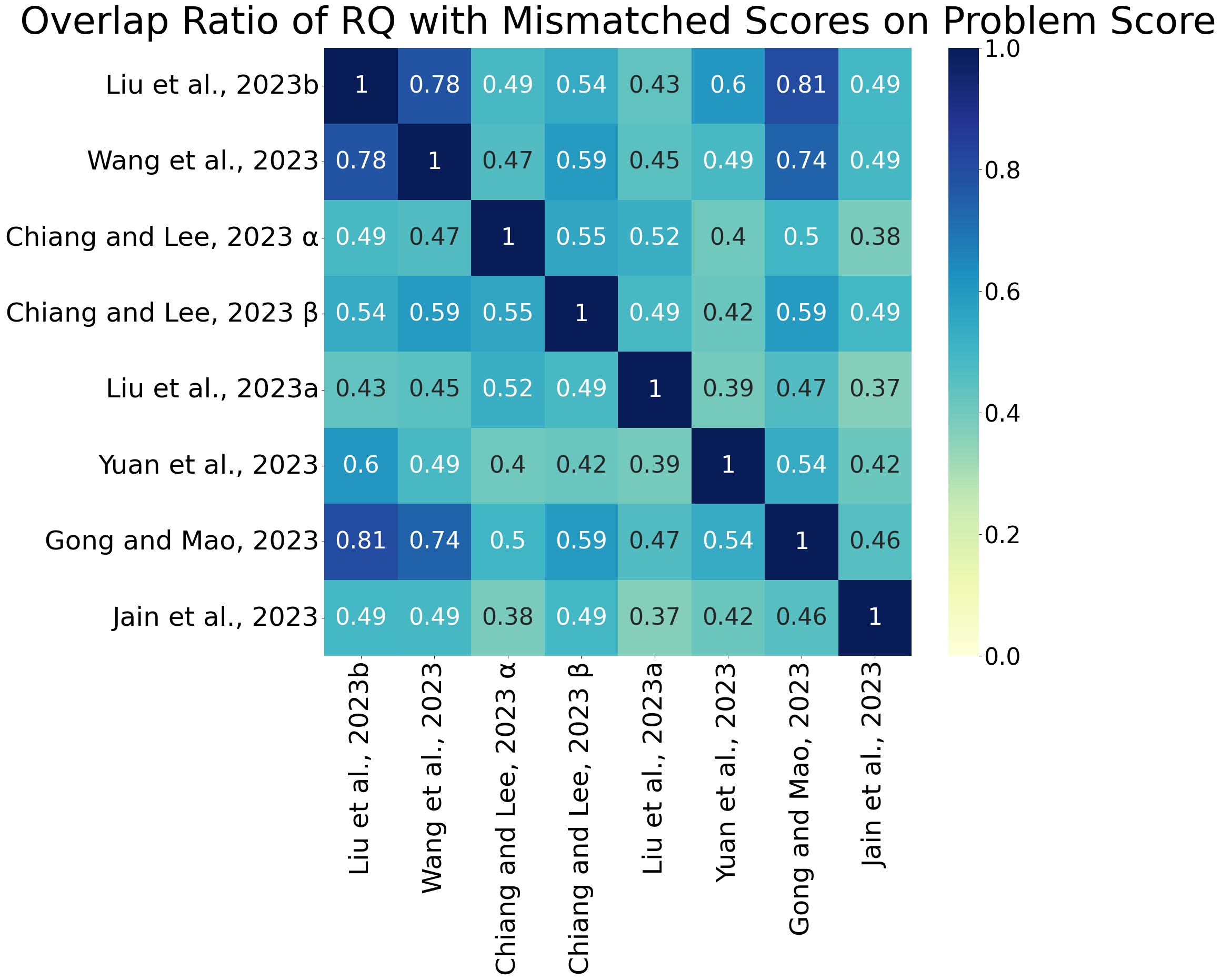}
\includegraphics[width=0.48\linewidth]{images/wrong_rq_overlap/method_score.png}
\includegraphics[width=0.48\linewidth]{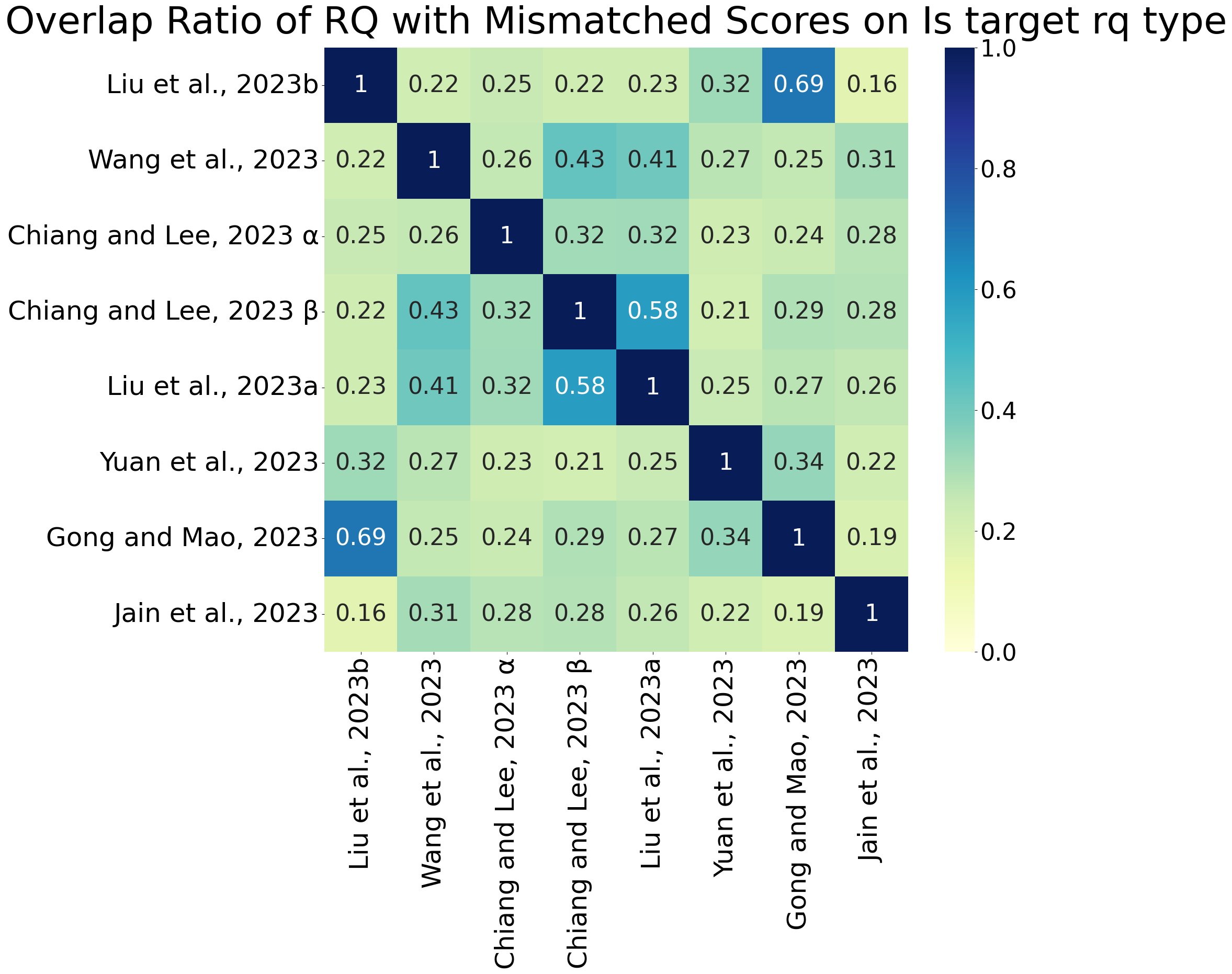}
\caption {Visualization of the overlap rate of RQ for which the evaluation values inferred by LLM did not match the GT between methods, categorized by Problem Score, Method Score, and Is target rq type, as a correlation.}
\label{fig:overlap-of-miss-format}
\end{figure*}

\Cref{fig:overlap-of-miss-format} visualizes the overlap rate between the sets of RQ for which the evaluation values inferred by LLM did not match the human-created GT for Is target rq type.

\subsubsection{Do more tokens lead to better performance?}
\label{sec:Do more tokens lead to better performance?}

As shown in the \Cref{sec:main-result}, the existing methods did not correlate well with human evaluations, indicating a need for designing better evaluation functions. To gain insights into what factors should be considered when designing such functions, we decided to investigate the properties of the relatively well-performing methods in our study.

Among the methods we employed, those by \citet{chiang-lee-2023-closer} and others performed relatively well, while \citet{Yuan2023BatchEvalTH}'s method performed poorly. One notable difference between these methods is the length of the input and output tokens. This observation led us to hypothesize that the number of input and output tokens might influence the model's performance. The basis for this hypothesis is, for example, \citet{Kojima2022zeroshotcot}'s research showing that adding a prompt encouraging multi-step reasoning to the input and performing multi-step reasoning at the output improved the model's performance. 

Based on this, we thought that methods that provide explanations for evaluation values, such as \citet{chiang-lee-2023-closer}, might have higher evaluation values than methods that simply output scores, such as \citet{Yuan2023BatchEvalTH}.

\begin{figure*}[t]
  \includegraphics[width=\linewidth]{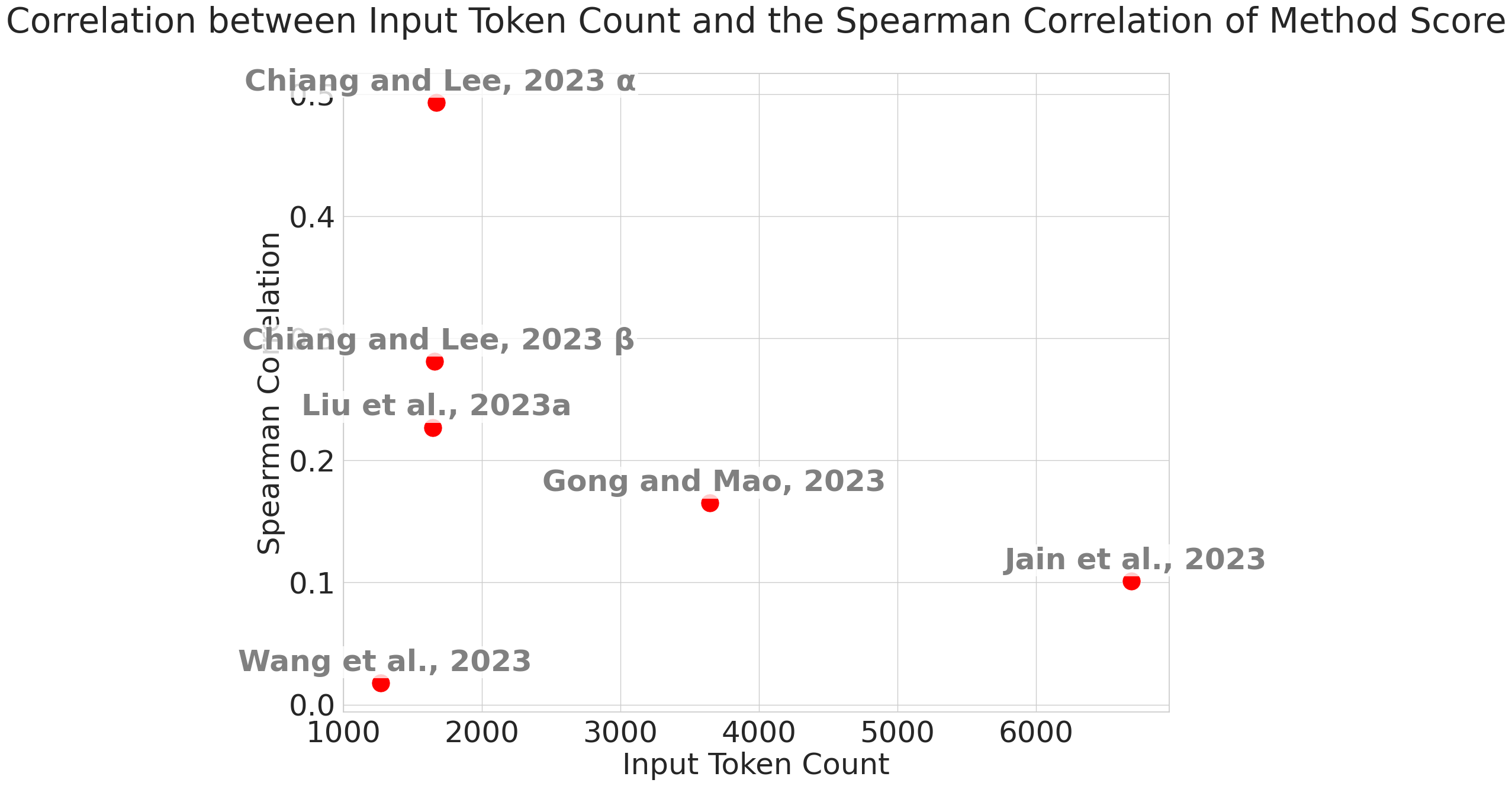}
  \includegraphics[width=\linewidth]{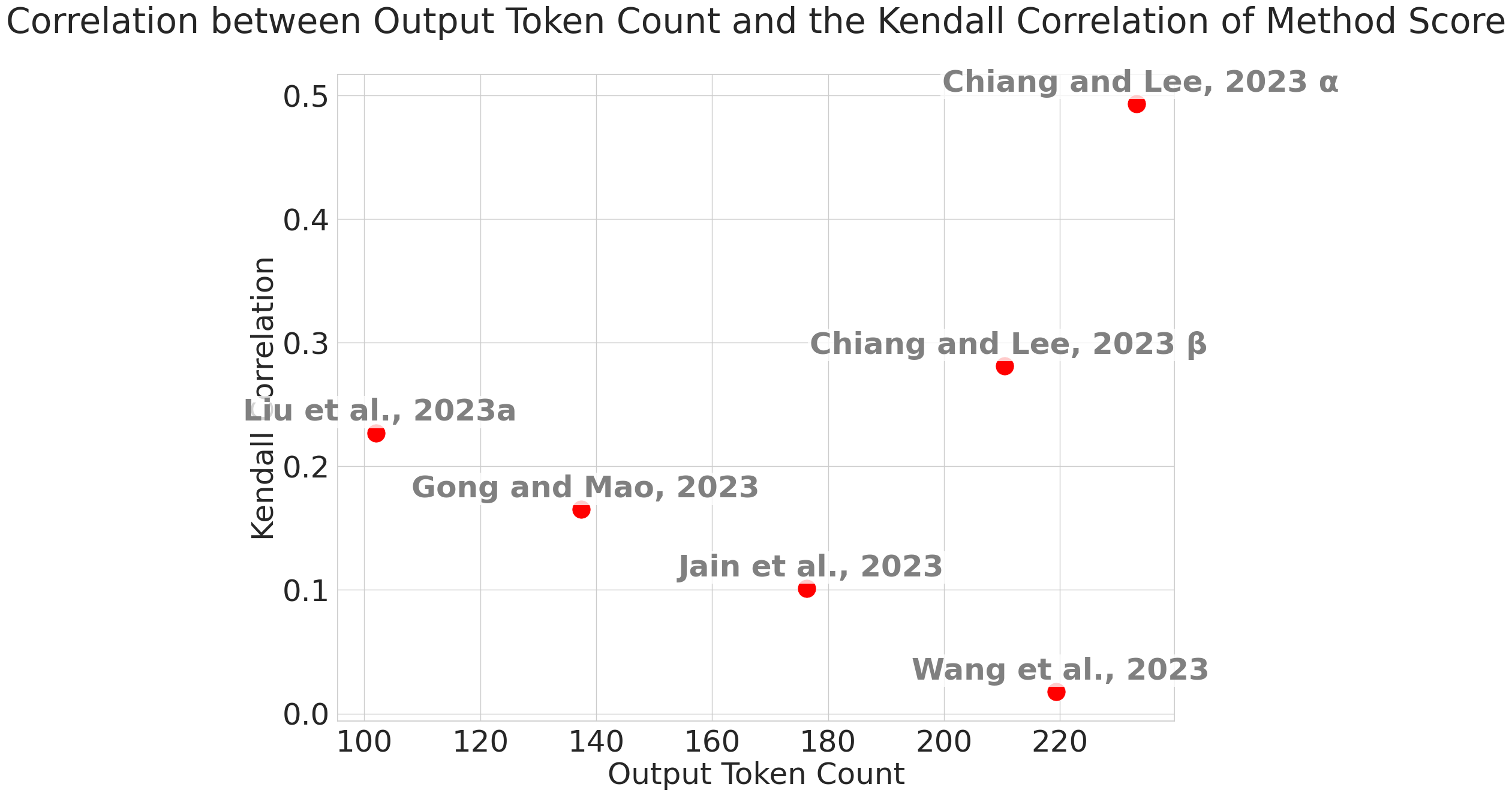}
  \caption {To investigate the relationship between the number of tokens and performance, we conducted an analysis using scatter plots. We plotted the number of tokens on the x-axis and the Spearman correlation coefficient, a performance indicator, on the y-axis, visualizing the positioning of each research method. The left figure shows the relationship between the number of input tokens and performance. The right figure shows the relationship between the number of output tokens and performance. There was a trend that the more output tokens there were, the higher the performance. However, since \citet{Yuan2023BatchEvalTH}'s method performs sampling in batches, it was difficult to calculate the number of tokens per sample. Therefore, \citet{Yuan2023BatchEvalTH}'s data is not included in this analysis.}
    \label{fig:experiments}
\end{figure*}

To confirm the relationship between the properties of these methods and the evaluation values, we visualized the performance against the number of tokens. In \Cref{fig:experiments}, we examined the relationship between the Spearman correlation coefficient of the Method Score, which yielded relatively good results, and the number of tokens. As a result, no clear correlation was found between the number of input or output tokens and the correlation coefficient of the Method Score. In other words, it is suggested that simply increasing the number of tokens does not yield automatic evaluation functions that are highly correlated with manual evaluation. It is highly likely that factors other than the number of tokens, such as the design of the evaluation function, are important for improving the performance of evaluation functions.

\subsubsection{Reproducibility of the Methods}

\paragraph{Impact of sample count on result variability}
\label{sec:appdx Impact of sample count on result variability}
 We conducted three trials each for \citet{wang-etal-2023-chatgpt} (single sampling) and \citet{chiang-lee-2023-closer} (20 samplings) and visualized the variability of Kendall correlation coefficients using violin plots.
\begin{figure}[t]
\includegraphics[width=\linewidth]{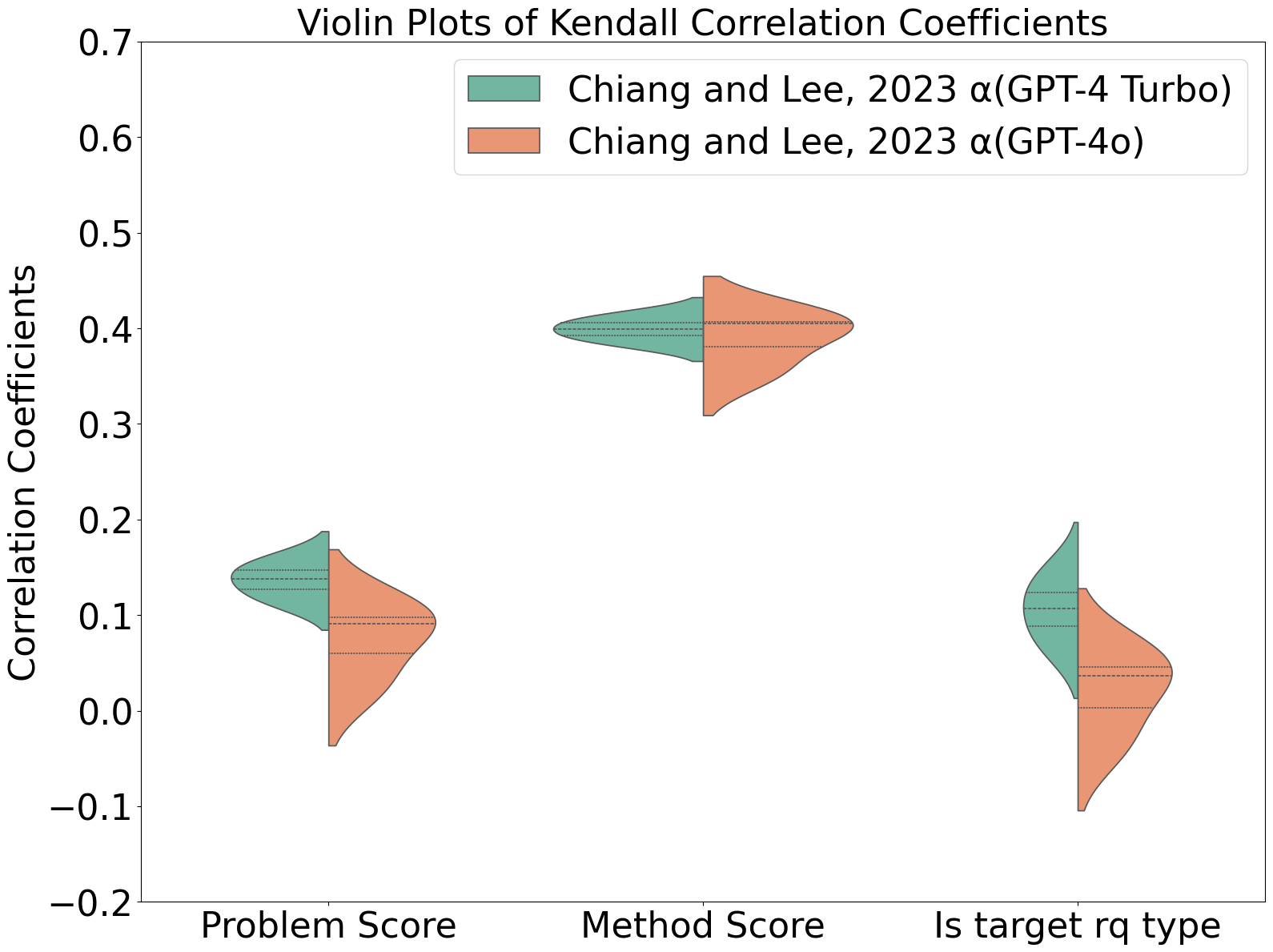}\hfill
\caption {Visualization of Kendall coefficients using violin plots, comparing \citet{wang-etal-2023-chatgpt} and \citet{chiang-lee-2023-closer} to confirm the variability due to differences in sample count.}
\label{fig:violinplot_diff_n_sample_kendal}
\end{figure}

\begin{figure}[t]
\includegraphics[width=\linewidth]{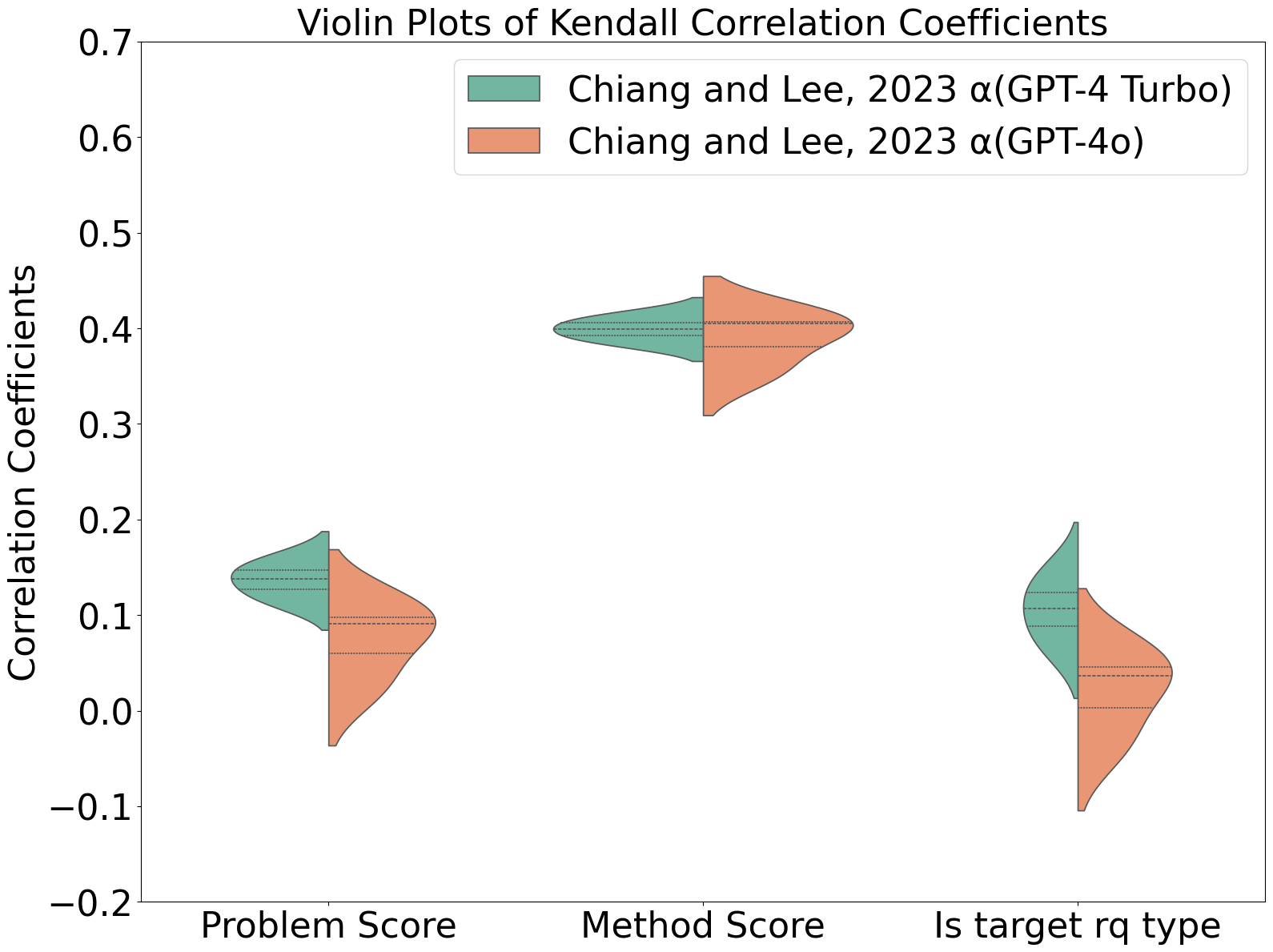}\hfill
\caption {Visualization of Kendall correlation coefficients using violin plots, comparing the analyze-rate of \citet{chiang-lee-2023-closer} to confirm the variability due to differences between \textit{gpt-4-turbo-2024-04-09} and \textit{gpt-4o-2024-05-13}.}
\label{fig:violinplot_diff_model_kendall}
\end{figure}

\paragraph{Variability of results due to model differences}
\label{sec:appdx Variability of results due to model differences}
We compared the variability of Kendall correlation coefficients between \textit{gpt-4-turbo-2024-04-09} and \textit{gpt-4o-2024-05-13} for the analyze-rate of \citet{chiang-lee-2023-closer} (best-performing method). While there was no significant difference in performance, \textit{gpt-4-turbo-2024-04-09} showed less variability (\Cref{fig:violinplot_diff_model_kendall}), suggesting that output results may fluctuate even if model performance does not vary significantly.

\subsubsection{Impact of Increasing the Number of Evaluation Procedure Steps on Performance}
\label{sec:Impact of Increasing the Number of Evaluation Procedure Steps on Performance}

The results \Cref{tab:result-extended-evaluation-procedure} show a slight improvement in performance, but the difference is small

\begin{table*}
\centering
\small
\setlength{\tabcolsep}{3.5pt}

\begin{tabular}{p{3.5cm}>{\centering\arraybackslash}p{1.2cm}>{\centering\arraybackslash}p{1.2cm}>{\centering\arraybackslash}p{1.2cm}>{\centering\arraybackslash}p{1.2cm}>{\centering\arraybackslash}p{1.2cm}>
{\centering\arraybackslash}p{1.2cm}>{\centering\arraybackslash}p{1.2cm}>{\centering\arraybackslash}p{1.2cm}}
\hline
{evaluation procedure steps} & \multicolumn{2}{c}{Problem Score} & \multicolumn{2}{c}{Method Score} & \multicolumn{2}{c}{Format Score}  \\

& $\rho$ & $\tau$ & $\rho$ & $\tau$ & $\rho$ & $\tau$  \\
\hline
default(approximately 5)  & 0.121 & 0.091 & 0.493& 0.405 & 0.067 & 0.055  \\
10 & \textbf{0.178} & \textbf{0.144} & \textbf{0.494}& 0.405 & 0.013 & 0.011  \\
30 & 0.153 & 0.124 & 0.485 & \textbf{0.410} & -0.058 & -0.055  \\
\hline
\end{tabular}
\caption{Performance confirmation was conducted by increasing the number of steps in the evaluation procedure. The default in the table refers to the original output, which is about 5 steps. In line with \Cref{tab:result}, the results in this table were also obtained using \textit{gpt-4-turbo-2024-04-09}.}
\label{tab:result-extended-evaluation-procedure}
\end{table*}

\subsubsection{Difference between estimated score and GT}
\label{sec:appdx overestimation of score}
\begin{figure*}[t]
\includegraphics[width=0.48\linewidth]{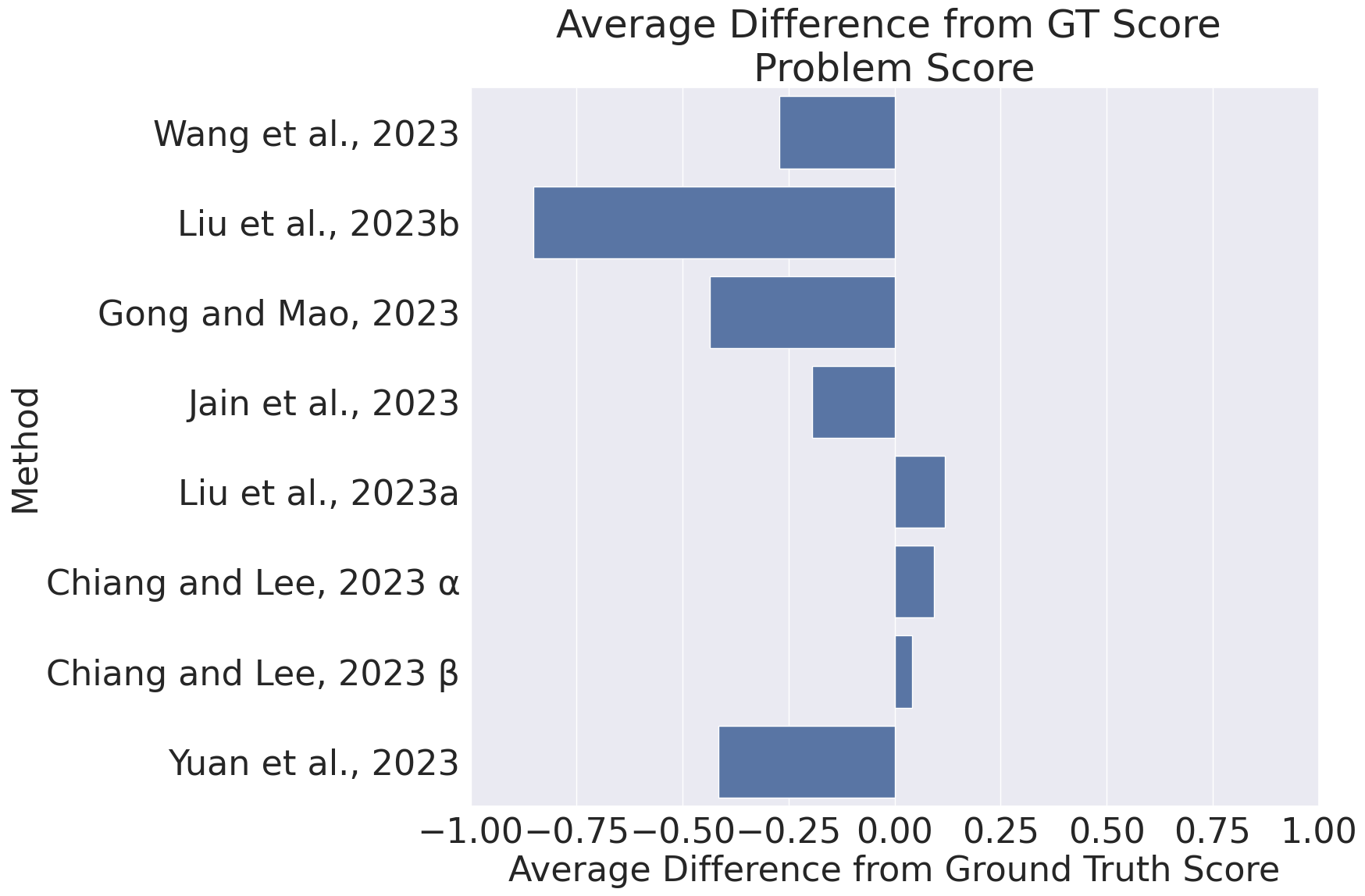}
\includegraphics[width=0.48\linewidth]{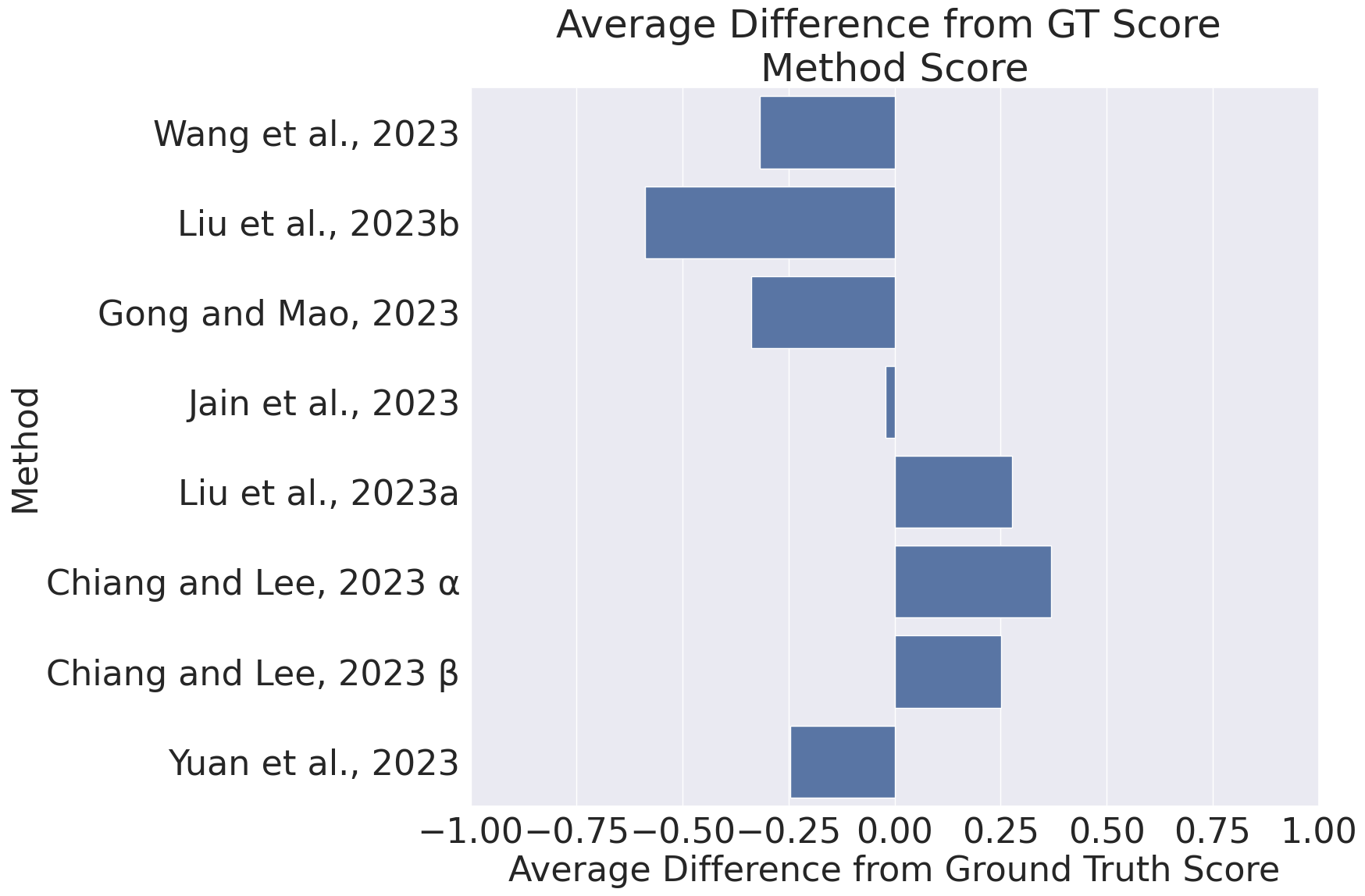}
\includegraphics[width=0.48\linewidth]{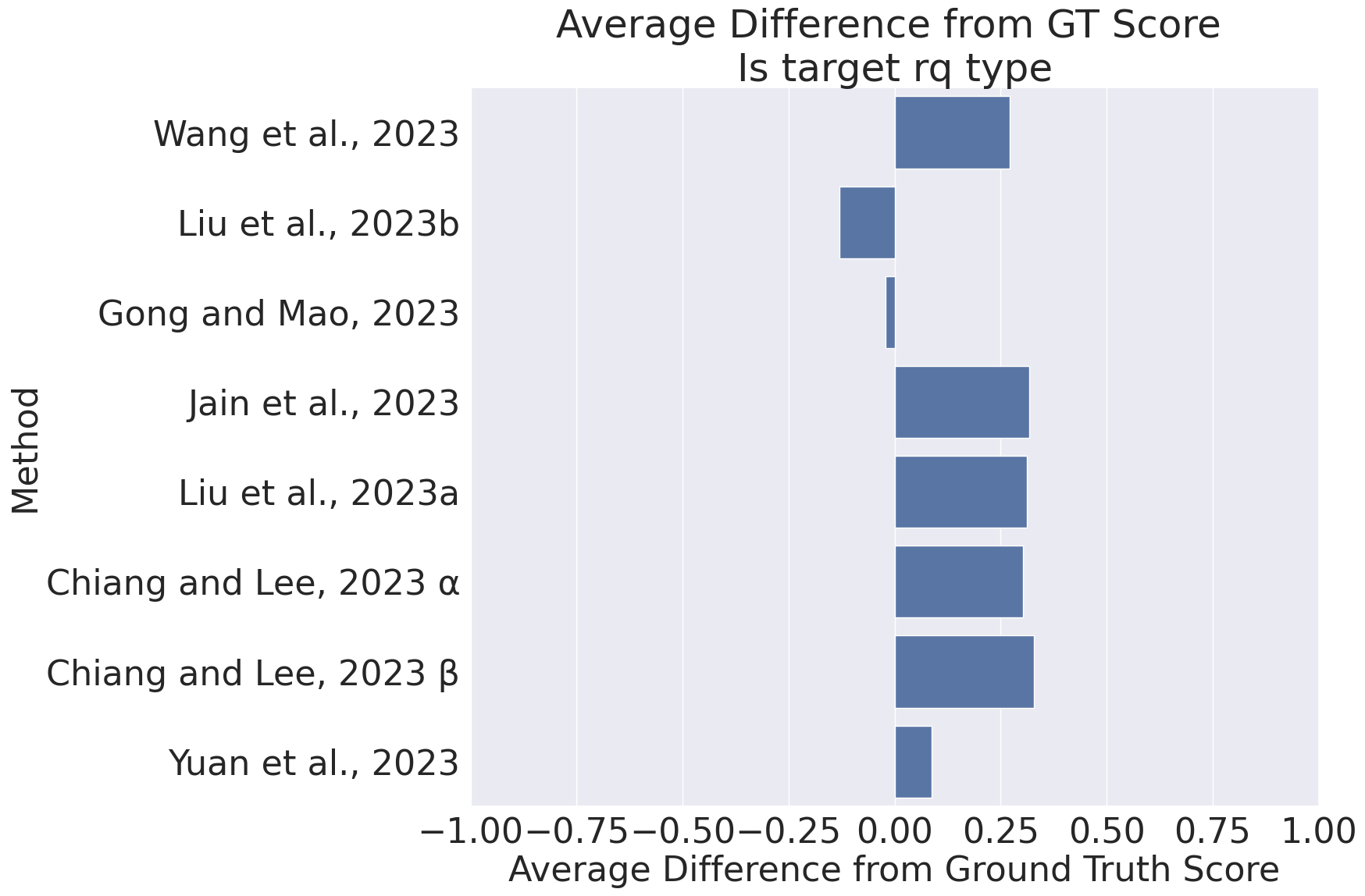}
\caption {Visualization of the difference between the estimated values of each method and the GT for each score}
\label{fig:overestimate-of-score}
\end{figure*}

In \Cref{fig:overestimate-of-score}, we visualize the difference between the estimated values of each method and the GT for each score.

\subsubsection{Learning the scoring patterns from the dataset}
\label{sec:appdx FT on Gemma}
As mentioned in \Cref{sec:Evaluation Functions}, the evaluation function used in this experiment is a type of evaluation function that involves trial and error with prompts using GPT-4. However, an alternative approach could be to fine-tune an LLM and learn the evaluation tendencies.
To this end, we fine-tuned the open-source LLM Gemma-7b by LoRA to see if it could better align with human ratings (GT). The experimental settings were LoRA rank of 8, alpha of 16, and 1 epoch. And A100 GPU were used, and the SFTTrainer from the Transformer Reinforcement Learning library was utilized. As shown in \Cref{tab:gemma-result}, the values are lower than those in \Cref{tab:result}, suggesting that it might be difficult to learn evaluation regularities by fine-tuning a model of around 7B parameters using LoRA. Furthermore, we investigated the impact on performance by varying the split ratio of the training data.

\begin{table*}
\centering
\small
\setlength{\tabcolsep}{3.5pt}

\begin{tabular}{p{2.5cm}>{\centering\arraybackslash}p{1.2cm}>{\centering\arraybackslash}p{1.2cm}>{\centering\arraybackslash}p{1.2cm}>{\centering\arraybackslash}p{1.2cm}>{\centering\arraybackslash}p{1.2cm}>
{\centering\arraybackslash}p{1.2cm}>{\centering\arraybackslash}p{1.2cm}>{\centering\arraybackslash}p{1.2cm}}
\hline
{Test data ratio} & \multicolumn{2}{c}{Problem Score} & \multicolumn{2}{c}{Method Score} & \multicolumn{2}{c}{Format Score}  \\

& $\rho$ & $\tau$ & $\rho$ & $\tau$ & $\rho$ & $\tau$  \\
\hline
10\% *& 0.084 & 0.080 & -0.061& -0.058 & -0.159 & -0.145  \\
40\%  & nan & nan & nan& nan & \textbf{0.171} & \textbf{0.164}  \\
70\%  & 0.094 & 0.089 & \textbf{0.136}& \textbf{0.128} & -0.165 & -0.158  \\
90\% & \textbf{0.125} & \textbf{0.120} & -0.184& -0.177 & nan & nan  \\
\hline
\end{tabular}
\caption{A list of performance for each test dataset. * indicates the same experimental settings as in \Cref{tab:result}, meaning that the results are comparable. As a result, many trials produced the same output values, leading to a large number of nan values in the table.}
\label{tab:gemma-result}
\end{table*}

\subsection{Actual Annotation Guidelines Used}
\label{sec:annotation-guideline}
\subsubsection{Introduction}
\paragraph{Purpose of this Task}
In this task, you will evaluate the accuracy of Research Questions (RQ) extracted by a Language Model (LLM) based on the abstract and introduction of research papers.
\paragraph{Types of RQ Covered in this Task}
RQ come in various forms, but for this task we will focus on papers with the following structure:
``Can a certain 'problem' be solved by a certain `method` ? ``
In other words, you will be assessing the accuracy of the RQ extracted by the LLM for papers that fit this specific template.
\paragraph{Utilization of the Evaluation}
The results of this evaluation can potentially be used to develop the following:
A model to classify whether a paper belongs to a particular RQ type.
A model to assess the validity of problems, challenges, or proposed methods extracted from a paper (by LLM or other means).
\subsubsection{Evaluation}
\paragraph{Evaluation Targets}
3 RQ extracted by the LLM.
\paragraph{Evaluation Procedure}
\label{paragraph : Evaluation Procedure of Problem Score}
The following is an evaluation procedure for the Problem Score.
\begin{itemize}
\item[1.] Carefully read the abstract and introduction of the paper
\item[2.] Extract the problem targeted by this research from the abstract and introduction of the paper
\item[3.] Confirm whether the problem targeted by this research is correct. For example, confirm whether the specific problem pointed out in the paper is correctly captured, rather than the abstract problem that the field is addressing
\item[4.] Based on the content confirmed in step 3, evaluate how accurately the extracted Research Question (RQ) captures the problem on a 3-point scale from 0 to 2. Refer to \Cref{tab:annotation-guideline}
\end{itemize}
\paragraph{Evaluation Items}
\begin{table*}
\centering
\small
\setlength{\tabcolsep}{3.5pt}
\begin{tabular}{|p{2cm}| p{8cm} |p{2cm} |p{2cm}|}
\hline
\textbf{Item} & \textbf{Description} & \textbf{Data Type} & \textbf{Content} \\
\hline
Problem Score & Determine how comprehensively the RQ extracted by the LLM captures the problems or challenges discussed in the target paper. & int & 0: Not mentioned, 1: Partially mentioned, 2: Comprehensively mentioned \\
\hline
Method Score & Determine how comprehensively the RQ extracted by the LLM captures the methods discussed in the target paper. If only the method name is mentioned, it is reasonable to consider it partially captured. & int &  0: Not mentioned, 1: Partially mentioned, 2: Comprehensively mentioned\\
\hline
Is target RQ type & Determine whether the RQ extracted by the LLM matches the following type:How are existing problems or challenges addressed by the proposed method? (= How effective is the proposed method in tackling existing problems or challenges?)Note that this is assessed independently of the Problem Score or Method Score.In other words, it simply determines if the type matches, regardless of the accuracy of the content. Also, general goals or metrics that can always be improved are not considered "currently identified problems or challenges". &int  &  0: Does not match, 1: Matches\\
\hline
\end{tabular}
\caption{Description of Each Evaluation Item}
\label{tab:annotation-guideline}
\end{table*}
\subsubsection{Notes}
Please evaluate the RQ in the order they appear from the top of the CSV file.

The "abstract" and "introduction" columns in the CSV file are generated through PDF parsing. Therefore, equations may not be accurately captured.

If the inaccuracy of equations makes it difficult to understand the paper's content, please skip evaluating that RQ.

\subsection{Utilization of AI Assistants in Research and Writing}
In this study, we mainly utilized AI assistants for creating Python scripts to conduct experiments and for checking spelling and typographical errors during paper writing.

\end{CJK}
\end{document}